\documentclass{article}
\usepackage[utf8]{inputenc}
\usepackage[T1]{fontenc}
\usepackage{textcomp}
\usepackage{authblk}

\usepackage{setspace}
\usepackage[margin=1.25in]{geometry}
\usepackage{graphicx}
\graphicspath{{./figure/}}
\usepackage{subcaption}
\usepackage{amsmath,amssymb,amsfonts}
\usepackage{amsthm}
\usepackage{mathrsfs}
\usepackage{xcolor}
\usepackage{booktabs}
\usepackage{tabularx}
\usepackage{array}
\usepackage{rotating}
\usepackage{placeins}
\usepackage{float}
\usepackage{threeparttable}
\usepackage[table]{xcolor}
\usepackage{tikz}
\usepackage{ragged2e}
\usepackage{multirow}
\usepackage{enumitem}

\usepackage[style=numeric-comp, sorting=none, maxbibnames=99, minbibnames=10, giveninits=true]{biblatex}
\usepackage[colorlinks=true, linkcolor=blue, citecolor=blue, urlcolor=blue, bookmarks=true]{hyperref}

\definecolor{zebragray}{HTML}{F7F5F1}
\definecolor{textgray}{HTML}{6B7084}
\definecolor{c1Bg}{HTML}{E8EDF5}      \definecolor{c1Bd}{HTML}{5E81AC}
\definecolor{c2Bg}{HTML}{EDF2E8}      \definecolor{c2Bd}{HTML}{A3BE8C}
\definecolor{c3Bg}{HTML}{FBF2ED}      \definecolor{c3Bd}{HTML}{D08770}
\definecolor{earlyBg}{HTML}{ECEFF4}   \definecolor{earlyBd}{HTML}{81A1C1}
\definecolor{interBg}{HTML}{F0F0F0}   \definecolor{interBd}{HTML}{9BA4B5}
\definecolor{lateBg}{HTML}{F0EAF8}    \definecolor{lateBd}{HTML}{9B7EC0}
\definecolor{v2xBg}{HTML}{E8EDF5}     \definecolor{v2xBd}{HTML}{5E81AC}
\definecolor{uxsBg}{HTML}{EDF2E8}     \definecolor{uxsBd}{HTML}{A3BE8C}
\definecolor{multiBg}{HTML}{FBF2ED}   \definecolor{multiBd}{HTML}{D08770}

\newcommand{\badge}[3]{%
	\tikz[baseline=(X.base)]\node
	(X)[draw=#2,fill=#1,rounded corners=1.6pt,
	inner xsep=1.5pt,inner ysep=1.0pt,line width=0.4pt,font=\tiny]{#3};%
}

\newcommand{\Cone}{\badge{c1Bg}{c1Bd}{C1}}
\newcommand{\Ctwo}{\badge{c2Bg}{c2Bd}{C2}}
\newcommand{\Cthree}{\badge{c3Bg}{c3Bd}{C3}}
\newcommand{\ConeTwo}{\Cone\;\Ctwo}
\newcommand{\ConeTwoThree}{\Cone\;\Ctwo\;\Cthree}
\newcommand{\ConeThree}{\Cone\;\Cthree}
\newcommand{\Cnone}{\badge{zebragray}{textgray}{none}}

\newcommand{\EarlyBadge}{\badge{earlyBg}{earlyBd}{Early}}
\newcommand{\InterBadge}{\badge{interBg}{interBd}{Intermed.}}
\newcommand{\LateBadge}{\badge{lateBg}{lateBd}{Late}}

\newcommand{\VTwoX}{\badge{v2xBg}{v2xBd}{V2X}}
\newcommand{\UxS}{\badge{uxsBg}{uxsBd}{UxS}}
\newcommand{\MultiDom}{\badge{multiBg}{multiBd}{Multi}}
\newcommand{\UxSLog}{\badge{uxsBg}{uxsBd}{UxS+Log.}}
\newcommand{\General}{\badge{zebragray}{textgray}{Gen.}}

\newcommand{\ygray}[1]{{\color{textgray}#1}}
\newcolumntype{L}[1]{>{\RaggedRight\arraybackslash}m{#1}}
\newcolumntype{C}[1]{>{\Centering\arraybackslash}m{#1}}
\definecolor{headergray}{RGB}{236,239,243}

\date{}

\begin{document}

\title{General Collaborative Intelligence: Architecting Cognition for Resilient Multi-Agent Ecosystems}

\author[1]{Lei Zhang}
\author[2]{Chun Ye}
\author[3]{Le Yang}
\author[4]{Zhaozhong Wang}
\author[5]{Deng-Ping Fan}
\author[2*]{Hang Dai}
\author[1*]{Binglu Wang}

\affil[1]{School of Astronautics, Northwestern Polytechnical University, Xi'an, China.}
\affil[2]{School of Computer Science, Wuhan University, Wuhan, China.}
\affil[3]{School of Electronics and Control Engineering, Chang'an University, Xi'an, China.}
\affil[4]{School of Automation, Northwestern Polytechnical University, Xi'an, China.}
\affil[5]{College of Computer Science, Nankai University, Tianjin, China.}
\affil[*]{Corresponding author: daihang@whu.edu.cn, wbl921129@gmail.com}

\maketitle

\begin{abstract}
Multi-agent unmanned systems are moving from isolated, ego-centric sensing toward collaborative intelligence, in which distributed agents exchange compact features to overcome a local observation trap that no single agent can escape: occlusions, finite sensor range, and environmental degradation. The field has matured across architectural, communication, embodied, resilience, and trust dimensions, yet existing surveys examine these dimensions in isolation and rarely expose the dependencies among them. This review offers a unified synthesis through two complementary lenses. The first is a five-dimensional taxonomy spanning collaboration stage, communication paradigm, fusion architecture, learning strategy, and application domain. The second is a set of three cognitive synergy conditions, Semantic Disambiguation, Pragmatic Information Exchange, and Proactive Informational Foraging, that turn the notion of cognitive synergy into operational criteria. Across these lenses we survey collaboration architectures and topologies, neural-communication co-design that treats the channel as a differentiable pipeline component, embodied action-perception loops via multi-agent reinforcement learning, and resilience mechanisms for synchronization, uncertainty quantification, and label-efficient learning. We then map these advances onto four operational domains, V2X, unmanned aerial, industrial logistics, and smart cities, and onto the safety-privacy-utility triad. To counter benchmark saturation and evaluation fragmentation, we propose GCI-Bench, a five-pillar scoring protocol with a maturity model that makes the trade-offs of collaborative methods comparable across studies. A critical reflection on reproducibility, the sim-to-real gulf, and conditions under which collaboration degrades performance identifies open challenges and charts directions toward general collaborative intelligence under real-world uncertainty.
\end{abstract}


\section{Introduction}\label{sec:intro}

\begin{figure}[t]
\centering
\includegraphics[width=\linewidth]{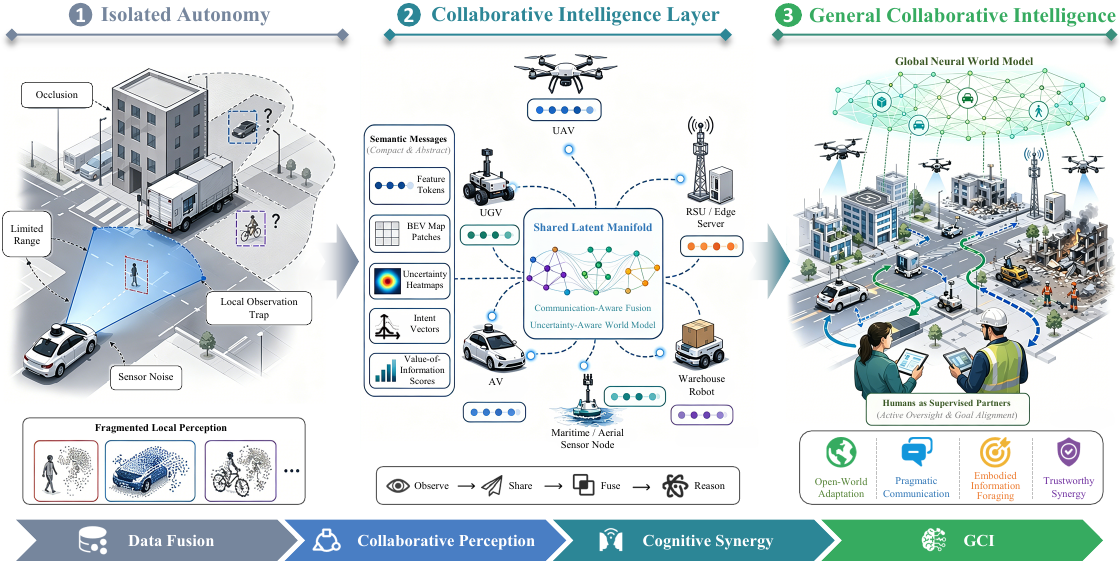}
\caption{\textbf{Vision of general collaborative intelligence for multi-agent unmanned systems.} The transition from isolated autonomy to GCI: heterogeneous agents exchange compact feature tokens through communication-aware fusion, evolving from deterministic data fusion toward cognitive synergy in a distributed ecosystem supporting open-world adaptation, pragmatic communication, embodied information foraging, and trustworthy synergy.}
\label{fig:gci_vision}
\end{figure}

Multi-agent unmanned systems operating across terrestrial, aerial, and maritime domains are undergoing a fundamental transition: from isolated, ego-centric sensing toward networked intelligence~\cite{goldberg2019robots, dorigo2020reflections, huang2025vehicle, billard2025roadmap, nitti2025collective, strobel2026foundation}. Single-agent perception, even when powered by large deep-learning models and high-fidelity sensors, remains bound by the local observation trap: physical occlusions, finite sensor range, and hardware-specific noise. These constraints impose an informational ceiling no single agent can shatter. Collaborative intelligence (CI) removes this ceiling by pooling distributed sensory, temporal, and computational resources into a shared world model, enabling a collective to execute missions that are physically and informationally impossible for any isolated agent. Figure~\ref{fig:gci_vision} illustrates this transition from isolated autonomy to the broader vision of general collaborative intelligence (GCI).

Historically, the field centered on deterministic data fusion, concerned primarily with rigid geometric alignment of raw sensory inputs, as surveyed in the multi-robot and SLAM literature~\cite{chen2023overview}. Since approximately 2020, however, the field has shifted toward \textbf{cognitive synergy}~\cite{zhuge2025mindstorms}, a collaborative state in which agents act as pragmatic informational gatekeepers of a shared semantic manifold, not as passive data broadcasters. This review treats GCI as a forward-looking systems target grounded in measurable capabilities; it does not claim that present systems already possess general intelligence, only that the conditions for cross-domain, task-relevant shared world modeling under heterogeneous sensing, constrained communication, agent churn, and safety and privacy requirements are beginning to be systematically characterized. Four converging trends drive this evolution: from geometric redundancy toward neural semantic consensus, from broadcast-driven sensing toward pragmatic informational foraging, from passive observation toward proactive cyber-physical acquisition, and from task-specific alignment toward GCI. Each trend is examined in Section~\ref{sec:arch}.

Progress notwithstanding, the transition from benchmarks to deployment faces fundamental obstacles. Recent surveys~\cite{wan2026systematic, song2025wireless, huang2025vehicle, yazgan2024survey, lyu2025survey} consistently identify a communication gap exceeding two orders of magnitude between raw sensor data rates and practical C-V2X throughput, agent heterogeneity across sensors, models, and protocols that breaks plug-and-play interoperability, localization errors and temporal asynchrony that produce ghosting artifacts, limited scalability beyond a handful of agents owing to \(O(N^2)\) fusion costs, online robustness challenges documented by end-to-end cooperative driving competitions~\cite{hao2025research}, and under-addressed privacy and trust risks. These obstacles, analyzed in Section~\ref{sec:open}, set the research agenda for the coming decade.

These challenges have prompted a wave of surveys, each distinct in scope. Wan et al.~\cite{wan2026systematic} provide a PRISMA-guided systematic review from a computer-vision standpoint. Song et al.~\cite{song2025wireless} reframe wireless communication as an ``information sensor.'' Huang et al.~\cite{huang2025vehicle} survey V2X cooperative perception across datasets, fusion methods, and privacy. Yazgan et al.~\cite{yazgan2024survey} examine intermediate-fusion methods through the lens of real-world robustness, and Lyu et al.~\cite{lyu2025survey} focus on redundancy mitigation. Among earlier studies, Han et al.~\cite{han2023collaborative} established the cooperative perception landscape of methods, datasets, and challenges, while Bai et al.~\cite{bai2024survey} frame the field from heterogeneous singletons toward hierarchical cooperation. Complementary reviews cover robotic-fleet perception~\cite{singh2024multi} and multi-robot perception, planning, and collaboration~\cite{chen2025survey}. An experimental study of V2X cooperative perception~\cite{hawlader2024cooperative} and a broader survey on causal inference for robust deep learning~\cite{research0467} further enrich the landscape.

Despite these contributions, four gaps motivate the present review. First, with the partial exception of Wan et al.~\cite{wan2026systematic} and Bai et al.~\cite{bai2024survey}, existing surveys concentrate on vehicle-to-everything (V2X) systems, leaving aerial-ground heterogeneous systems (UxS), industrial warehouse logistics, and smart-city infrastructure under-represented. Second, communication is treated predominantly as an external constraint rather than as a differentiable, co-optimized pipeline component; the emerging paradigm of neural-communication co-design, including Deep JSCC, information-bottleneck-guided feature pruning, and cross-layer MAC-PHY orchestration, therefore remains unsystematized. Third, no prior survey offers a unified operational framework, grounded in formal definitions and operational conditions, that ties architectural choices, learning strategies, and domain deployments to common evaluative principles. Fourth, while individual challenges such as localization error, latency, and heterogeneity are cataloged, their structural drivers (i.e., benchmark saturation, the reproducibility gap, the sim-to-real performance gulf, and failure modes where collaboration degrades performance) have not been systematically examined. The present review addresses these structural dimensions in a dedicated critical reflection (Section~\ref{sec:open}). Table~\ref{tab:related_reviews} provides a systematic comparison with representative prior surveys across four groups and nine dimensions of coverage.

\begin{table}[t]
\centering
\begin{threeparttable}
	\caption{\textbf{Systematic comparison of this review with representative prior surveys.} A filled circle ($\bullet$) denotes comprehensive coverage; an open circle ($\circ$) denotes partial coverage; a dash (--) indicates the dimension is not addressed.}
	\label{tab:related_reviews}
	\renewcommand{\arraystretch}{1.50}
	\tiny
	\setlength{\tabcolsep}{11pt}
	\renewcommand{\tabularxcolumn}[1]{m{#1}}
	\sffamily
	\begin{tabular}{c c *{9}{c}}
		\toprule
		\multirow{2}{*}{} & \multirow{2}{*}{} & \multicolumn{2}{c}{\textbf{Application Domain}} & \multicolumn{2}{c}{\textbf{Technical Core}} & \multicolumn{2}{c}{\textbf{Tasks \& Robustness}} & \multicolumn{3}{c}{\textbf{System \& Reflection}} \\
		\cmidrule(lr){3-4} \cmidrule(lr){5-6} \cmidrule(lr){7-8} \cmidrule(lr){9-11}
		\multirow{-2}{*}{\textbf{Ref.}} & \multirow{-2}{*}{\textbf{Year}} & \textbf{V2X} & \textbf{UxS/ISC} & \textbf{Co-Des.} & \textbf{Formal.} & \textbf{Tasks} & \textbf{Chall.} & \textbf{Embod.} & \textbf{Trust} & \textbf{Refl.} \\
		\midrule
		Han et al.~\cite{han2023collaborative} & \ygray{2023} & $\bullet$ & -- & -- & -- & $\bullet$ & $\circ$ & -- & -- & -- \\
		\rowcolor{gray!6} Yazgan et al.~\cite{yazgan2024survey} & \ygray{2024} & $\bullet$ & -- & $\circ$ & -- & $\bullet$ & $\circ$ & -- & $\circ$ & $\circ$ \\
		Bai et al.~\cite{bai2024survey} & \ygray{2024} & $\bullet$ & $\circ$ & -- & -- & $\circ$ & -- & -- & -- & -- \\
		\rowcolor{gray!6} Song et al.~\cite{song2025wireless} & \ygray{2025} & $\bullet$ & -- & $\bullet$ & -- & $\circ$ & $\bullet$ & -- & -- & -- \\
		Huang et al.~\cite{huang2025vehicle} & \ygray{2025} & $\bullet$ & -- & $\circ$ & -- & $\circ$ & $\bullet$ & -- & $\circ$ & -- \\
		\rowcolor{gray!6} Lyu et al.~\cite{lyu2025survey} & \ygray{2025} & $\bullet$ & -- & $\circ$ & -- & $\circ$ & $\circ$ & -- & -- & -- \\
		Wan et al.~\cite{wan2026systematic} & \ygray{2026} & $\bullet$ & -- & $\circ$ & -- & $\bullet$ & $\bullet$ & -- & $\circ$ & $\circ$ \\
		\rowcolor{gray!6} \textbf{This review} & \ygray{2026} & $\bullet$ & $\bullet$ & $\bullet$ & $\bullet$ & $\circ$ & $\bullet$ & $\bullet$ & $\circ$ & $\bullet$ \\
		\bottomrule
	\end{tabular}%
	\smallskip
	\parbox{\linewidth}{\footnotesize \textit{Abbreviations:} Co-Des. = Communication Co-Design; Formal. = Formal Mathematical Framework; Tasks = Perception Tasks (detection, tracking, segmentation, prediction); Chall. = Practical Challenges (bandwidth, latency, localization error, heterogeneity, domain shift, etc.); Embod. = Embodied Synergy (MARL, active perception, sim-to-real); Trust = Trust \& Privacy; Refl. = Critical Reflection \& Reproducibility; UxS/ISC = Unmanned Systems, Industrial Logistics, Smart Cities.}
\end{threeparttable}
\end{table}

Against this background, the present review adopts an analytical lens introduced in Section~\ref{sec:foundations} and applied throughout the paper. This lens combines a \textbf{five-dimensional taxonomy} (collaboration stage, communication paradigm, fusion architecture, learning strategy, application domain) with three \textbf{cognitive synergy conditions}: Semantic Disambiguation (C1), Pragmatic Information Exchange (C2), and Proactive Informational Foraging (C3). Our main contributions are:
\begin{itemize}
\item We operationalize cognitive synergy through the three operational conditions C1--C3, which jointly require that a collaborative system resolve entity identity beyond any single agent's view, transmit only decision-relevant informational surprise, and actively reconfigure its sensing topology to reduce residual uncertainty. These conditions serve as the evaluative lens for the five-dimensional taxonomy and for the GCI-Bench scoring protocol.
\item Unlike prior surveys that focus on V2X, we synthesize advances across four operational domains (V2X, UxS, industrial logistics, and smart cities) and five technical dimensions (architectures, communication co-design, embodied synergy, resilience, and trust), while explicitly distinguishing mature benchmark-supported areas from emerging application domains.
\item Beyond cataloging progress, we provide a structured critique of benchmark saturation, the reproducibility gap, the sim-to-real gulf, under-reported failure modes, and scalability challenges, offering a counterweight to the positive-results bias prevalent in the literature.
\item We identify four research trajectories that chart the transition from task-specific coordination to general collaborative intelligence: collaborative foundation models, pragmatic game-theoretic communication, hardware-agnostic semantics, and proactive swarm embodiment.
\end{itemize}

\noindent\textbf{Review scope and methodology.}
This review follows a structured scoping procedure rather than a bibliometric meta-analysis. We searched IEEE Xplore, ACM Digital Library, SpringerLink, ScienceDirect, Web of Science, Google Scholar, and arXiv for literature published primarily between 2019 and May 2026, retaining papers that propose collaborative sensing, fusion, or communication methods for autonomous vehicles, unmanned systems, or multi-robot platforms; introduce relevant datasets, simulators, or benchmarks; address deployment constraints (latency, pose error, bandwidth, heterogeneity, robustness, privacy, safety); or provide a recent survey defining the state of the field. Single-agent perception, purely network-level V2X studies without perceptual tasks, and application reports without reusable technical insight were excluded.

The remainder of this paper is organized as follows. Section~\ref{sec:foundations} establishes the conceptual and mathematical foundations. Sections~\ref{sec:arch}--\ref{sec:resilience} examine four technical dimensions: collaboration architectures (stage and topology), neural-communication co-design, embodied synergy, and resilience. Sections~\ref{sec:domain}--\ref{sec:trust} map these advances onto application domains and the safety--privacy--utility triad. Section~\ref{sec:open} provides a structured critical reflection on current limitations and charts future directions. Section~\ref{sec:gci-bench} presents the GCI-Bench evaluation framework, and Section~\ref{sec:conclusion} concludes.

\section{Foundations of Collaborative Intelligence}\label{sec:foundations}

Before surveying the technical landscape, we establish the conceptual and mathematical foundations that unify the diverse methods discussed in subsequent sections. This section formalizes the fundamental tension between perceptual fidelity and communication cost that governs all collaborative paradigms, introduces an operational definition of cognitive synergy as the evaluative lens used throughout the review, and situates the field within a multi-dimensional taxonomy that organizes the remainder of the paper.

\subsection{The Rate--Fidelity Trade-off}

The structural organization of GCI is fundamentally governed by a multi-objective optimization problem: the tension between perceptual fidelity and the entropy of the communication substrate. Formally, let $N$ agents each observe a local sensory field $\mathbf{O}_i$, encode it into a latent representation $\mathbf{Z}_i = f_i(\mathbf{O}_i)$ with transmission rate $R(\mathbf{Z}_i) \triangleq I(\mathbf{O}_i; \mathbf{Z}_i)$ (mutual information, information-bottleneck convention), and contribute to a fused estimate $\hat{\mathbf{Y}} = g(\mathbf{Z}_1, \ldots, \mathbf{Z}_N)$ of the global world state $\mathbf{Y}$. The collaborative architecture design problem is:
\begin{equation}
\max_{f_i, g} \; I(\hat{\mathbf{Y}}; \mathbf{Y}) \quad \text{subject to} \quad \sum_{i=1}^{N} R(\mathbf{Z}_i) \leq B,
\label{eq:arch_ib}
\end{equation}
where $I(\cdot;\cdot)$ denotes mutual information and $B$ is the total communication budget. This formulation captures the core architectural trade-off: increasing the information content of transmitted features improves perceptual fidelity at the cost of communication load. The three canonical paradigms (early, intermediate, and late collaboration) correspond to different operating points along this rate--fidelity frontier. Early collaboration maximizes $I(\hat{\mathbf{Y}}; \mathbf{Y})$ by setting $\mathbf{Z}_i \approx \mathbf{O}_i$ (raw data), at the expense of maximal $R(\mathbf{Z}_i)$; late collaboration minimizes $R(\mathbf{Z}_i)$ by transmitting only sparse metadata, but incurs a substantial drop in mutual information due to the hard thresholding of local decisions; intermediate collaboration seeks a Pareto-optimal balance by learning a compressed neural code $\mathbf{Z}_i$ that preserves task-relevant semantics while discarding redundancy~\cite{chen2019cooper, hu2022where2comm, hu2026pragmatic}.

To avoid notational drift across the technical sections that follow, we fix the shared symbol vocabulary introduced above and reuse it throughout the review: $N$ denotes the number of collaborating agents; $\mathbf{O}_i$ the local observation of agent $i$; $\mathbf{Z}_i = f_i(\mathbf{O}_i)$ its transmitted latent representation; $R(\mathbf{Z}_i) \triangleq I(\mathbf{O}_i; \mathbf{Z}_i)$ the information-theoretic transmission rate; $\hat{\mathbf{Y}} = g(\mathbf{Z}_1, \ldots, \mathbf{Z}_N)$ the fused collective estimate; and $\mathbf{Y}$ the underlying global world state. Operational communication cost (bandwidth plus latency, as measured in practice) is denoted $C$ and is distinguished from the information-theoretic rate $R$; movement cost in embodied settings is written $C_{\text{move}}$. Section-specific symbols (e.g., model parameters $\boldsymbol{\theta}$, agent poses $\mathbf{p}_i$) are introduced locally where they arise.

This rate--fidelity framework is not confined to architectural choices. It directly underpins the collaborative information bottleneck formulation (Eq.~\ref{eq:ib_collaborative}) and the joint perception--communication Lagrangian (Eq.~\ref{eq:unified_lagrangian}) developed in Section~\ref{sec:comm}, establishing a unified mathematical language that spans the architectural, communication, and learning dimensions of GCI.

\subsection{Operationalization of Cognitive Synergy}

\begin{table}[t]
\centering
\begin{threeparttable}
	\caption{\textbf{Operational criteria for cognitive synergy.} The three conditions translate the conceptual claim of GCI into observable requirements that can be evaluated experimentally.}
	\label{tab:synergy_operational}
	\renewcommand{\arraystretch}{1.40}
	\tiny
	\setlength{\tabcolsep}{10pt}
	\renewcommand{\tabularxcolumn}[1]{m{#1}}
	\sffamily
	
	\begin{tabularx}{\textwidth}{
			>{\hsize=1.0\hsize\raggedright\arraybackslash}X 
			>{\hsize=1.1\hsize\raggedright\arraybackslash}X 
			>{\hsize=0.9\hsize\raggedright\arraybackslash}X 
			>{\hsize=1.0\hsize\raggedright\arraybackslash}X
		}
		\toprule
		\centering\textbf{Condition} & 
		\centering\textbf{Observable Criterion} & 
		\centering\textbf{Typical Technical Realization} & 
		\centering\arraybackslash\textbf{Operational Failure Test (metric; benchmark)} \\
		\midrule
		
		\Cone\ Semantic Disambiguation & 
		Resolve entity identity, spatial location, and trajectory intent beyond any single agent's line-of-sight & 
		Cross-attention Transformers~\cite{xu2022v2x, xu2023cobevt}; generative map priors (CoGMP)~\cite{fu2025generative} &
		On a dynamic-occlusion split of OPV2V~\cite{xu2022opv2v}: identity-switch or class-misattribution rate above a set threshold (e.g., $>$5\%) as targets cross multi-agent occlusions \\
		
		\cellcolor{gray!6} \Ctwo\ Pragmatic Exchange & 
		\cellcolor{gray!6} Bandwidth scales inversely with spatial confidence; high-entropy, high-surprise regions transmitted first & 
		\cellcolor{gray!6} Extended information bottleneck (InfoCom~\cite{wei2026infocom}); spatial confidence masking (Where2comm~\cite{hu2022where2comm}) &
		\cellcolor{gray!6}On OPV2V~\cite{xu2022opv2v} or V2X-Sim~\cite{li2022v2x}: communication volume scaling near-linearly with agent density (log--log slope $\gtrsim 0.8$), or a non-negligible share of redundant background bytes \\
		
		\Cthree\ Proactive Foraging & 
		Trajectory adjustments or sensor re-aiming triggered by localized state-uncertainty peaks & 
		Active perception loops; planning-integrated perception (Plan2comm~\cite{xie2024towards}, CooperDrive~\cite{qu2026cooperdrive}) &
		On an active-perception testbed (e.g., CSAOT~\cite{nguyen2025csaot}): no measurable trajectory or sensor reconfiguration, and hence no entropy reduction, in regions of high epistemic uncertainty \\
		\bottomrule
	\end{tabularx}
\end{threeparttable}
\end{table}

To ground the subsequent discussion, we define \textbf{cognitive synergy} operationally as a collaborative state characterized by three operationally necessary conditions; we treat these as design desiderata rather than a formal theorem, and note that each condition addresses a distinct failure mode that the others cannot compensate. 
\textbf{(C1) Semantic Disambiguation}: the collective resolves entity identity beyond any single agent's local view, grounding synergy in semantic understanding rather than mere spatial occupancy.
\textbf{(C2) Pragmatic Information Exchange}: agents transmit only decision-relevant surprise, ensuring communication carries information that reduces collective belief-state entropy.
\textbf{(C3) Proactive Informational Foraging}: the swarm reconfigures its physical or attentional topology to actively acquire information that resolves residual uncertainty.

When all three conditions hold, the swarm exhibits identity persistence, graceful degradation, and cross-modal invariance; systems satisfying any subset of the three conditions fall into regimes where synergy is incomplete, either lacking embodiment, semantic depth, or bandwidth discipline.

The three conditions hold only within an operating envelope, and recognizing this envelope is part of the concept's definition rather than an afterthought: if the conditions are meant to characterize cognitive synergy, then the conditions under which collaboration should \emph{not} be attempted belong to that definition. Collaboration should be withheld, or should fall back to single-agent operation, when any of the following obtains. (i) The end-to-end latency of encoding, transmission, and fusion exceeds the decision budget of the task, so shared features arrive too stale to help; this is a failure of C2's timeliness and is the dominant regime in time-critical scenarios such as highway emergency braking. (ii) The agents share a common blind spot or a homogeneous failure mode, so pooling reinforces a correlated error instead of disambiguating identity; this is a failure of C1. (iii) A compromised or miscalibrated agent injects corrupted features faster than fusion can reject them, so the collective estimate becomes worse than the best local estimate; this is the error-propagation failure. (iv) The communication cost of resolving residual uncertainty exceeds the value of the information gained, so C3's foraging consumes more resources than it returns. These four regimes, analyzed as the field's documented failure modes in Section~\ref{sec:open}, mark the boundary at which synergy degrades into anti-synergy, and a system operating beyond it satisfies none of the three conditions (denoted \Cnone\ below).

\subsection{A Multi-Dimensional Taxonomy of Collaborative Perception}

\begin{figure}[t]
\begin{center}
	\includegraphics[width=\linewidth]{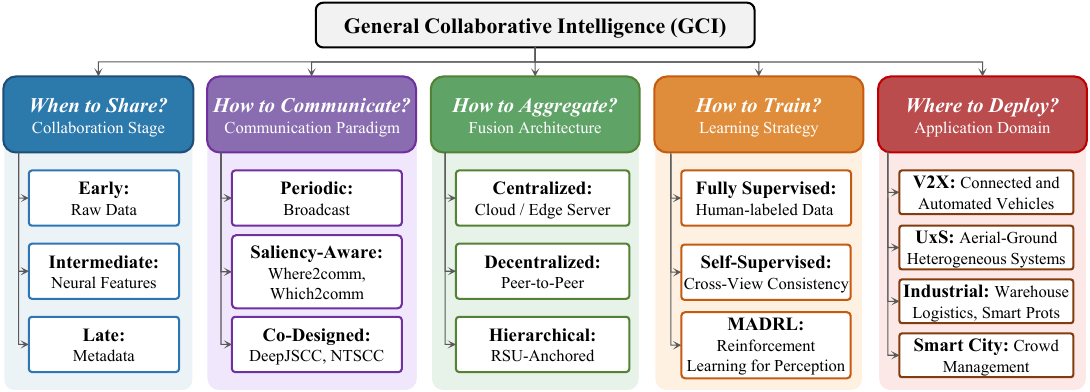}
\end{center}
\caption{\textbf{A multi-dimensional taxonomy of collaborative perception.} The taxonomy organizes the field along five orthogonal dimensions. \textbf{Dimension 1} (Collaboration Stage) determines the abstraction level at which information is exchanged, directly governing the rate--fidelity trade-off formalized in Eq.~\ref{eq:arch_ib}. \textbf{Dimension 2} (Communication Paradigm) spans the spectrum from periodic broadcasting to neural-communication co-design (Section~\ref{sec:comm}). \textbf{Dimension 3} (Fusion Architecture) specifies the topological structure of information aggregation. \textbf{Dimension 4} (Learning Strategy) determines how the collaborative model acquires its parameters. \textbf{Dimension 5} (Application Domain) captures the operational context that drives domain-specific design choices (Section~\ref{sec:domain}).}
\label{fig:taxonomy}
\end{figure}

Existing surveys typically classify collaborative perception systems along a single axis, most commonly the collaboration stage (early, intermediate, late) or the application domain. This one-dimensional view, however, obscures critical cross-cutting concerns: a late-fusion system deployed in a decentralized V2X topology faces fundamentally different design trade-offs than a late-fusion system in a centralized warehouse setting, yet both share the same ``late collaboration'' label under a stage-only taxonomy. A multi-dimensional taxonomy is therefore necessary to capture the design space accurately: each dimension represents an independent axis of variation, and the full characterization of a collaborative system requires specifying its position along all five axes simultaneously.

Figure~\ref{fig:taxonomy} visualizes this taxonomy and serves as the organizing framework for the remainder of the paper. Section~\ref{sec:gci-bench} operationalizes this taxonomy into a quantitative multi-pillar scoring protocol (GCI-Bench). Dimension~1 (Collaboration Stage) and Dimension~3 (Fusion Architecture) are examined together in Section~\ref{sec:arch}, where we emphasize their orthogonality: any collaboration stage can in principle be combined with any topology, and conflating them is a common source of confusion in the literature. Dimension~2 (Communication Paradigm) is covered in Section~\ref{sec:comm}. Dimension~4 (Learning Strategy) cuts across Sections~\ref{sec:embodied} and~\ref{sec:resilience}. Dimension~5 (Application Domain) is the subject of Section~\ref{sec:domain}.

To demonstrate the five dimensions of Figure~\ref{fig:taxonomy}, each subsequent section opens with a taxonomy table that maps the methods discussed therein onto all five axes and the three cognitive synergy conditions (see Table~\ref{tab:synergy_operational}). Synergy pillars are denoted as \Cone, \ConeTwo, or \ConeTwoThree\  according to which conditions each method satisfies; failure regimes that satisfy none of the conditions are marked \Cnone. These badges are assigned by a single adjudication rule: a method is credited with a condition only when its core technical contribution directly implements the corresponding observable criterion of Table~\ref{tab:synergy_operational}, rather than benefiting from that condition incidentally; C1 is credited in a strong form when fusion resolves entity identity at the feature level and in a weak form when identity is resolved only through cross-view decision association. Applying this rule, we re-audited every assignment in Tables~\ref{tab:taxonomy_arch}--\ref{tab:taxonomy_open}: DiscoNet, for example, is credited with \Cone\ alone, since its edge distillation compresses the collaboration graph rather than gating decision-relevant surprise (C2), whereas late fusion retains a weak-form \Cone\ through metadata association across agent decisions. Table~\ref{tab:taxonomy_arch} opens Section~\ref{sec:arch} by indexing 13 representative architecture methods; analogous tables appear at the start of Sections~\ref{sec:comm}--\ref{sec:open} (Tables~\ref{tab:taxonomy_comm}--\ref{tab:taxonomy_open}), collectively forming a distributed taxonomy index that covers the methodological landscape surveyed in this review.

\section{Collaboration Architectures and Topologies}\label{sec:arch}

\begin{table}[t]
\centering
\begin{threeparttable}
	\caption{\textbf{Taxonomy of collaboration architectures and cognitive synergy conditions.} Methods are mapped onto the proposed five-dimensional framework.}
	\label{tab:taxonomy_arch}
	\renewcommand{\arraystretch}{1.5}
	\tiny
	\setlength{\tabcolsep}{4pt}
	\renewcommand{\tabularxcolumn}[1]{m{#1}}
	\sffamily
	\begin{tabularx}{\textwidth}{
			>{\raggedright\arraybackslash\hsize=1.3\hsize}X
			>{\centering\arraybackslash\hsize=0.7\hsize}X 
			>{\centering\arraybackslash\hsize=0.8\hsize}X 
			>{\raggedright\arraybackslash\hsize=1.4\hsize}X
			>{\centering\arraybackslash\hsize=0.9\hsize}X 
			>{\raggedright\arraybackslash\hsize=1.5\hsize}X
			>{\centering\arraybackslash\hsize=0.7\hsize}X 
			>{\centering\arraybackslash\hsize=0.7\hsize}X}
		\toprule
		\textbf{Method} & \textbf{Year} & \textbf{Stage} & \textbf{Comm. Paradigm} & \textbf{Architecture} & \textbf{Learning Strategy} & \textbf{Application} & \textbf{Synergy} \\
		\midrule
		Cooper~\cite{chen2019cooper} & \ygray{2019} & \EarlyBadge & Broadcast & Centr. & Supervised & \VTwoX & \Cone \\
		\rowcolor{gray!6} Late Fusion~\cite{yu2022dair} & \ygray{2022} & \LateBadge & Metadata-only & Decentr. & Supervised & \VTwoX & \Cone \\
		V2VNet~\cite{wang2020v2vnet} & \ygray{2020} & \InterBadge & Broadcast (GNN) & Decentr. & Supervised & \VTwoX & \Cone \\
		\rowcolor{gray!6} DiscoNet~\cite{li2021learning} & \ygray{2021} & \InterBadge & KD-driven & Decentr. & Supervised + Distill. & \VTwoX & \Cone \\
		V2X-ViT~\cite{xu2022v2x} & \ygray{2022} & \InterBadge & Broadcast (Transf.) & Centr. & Supervised & \VTwoX & \Cone \\
		\rowcolor{gray!6} Where2comm~\cite{hu2022where2comm} & \ygray{2022} & \InterBadge & Saliency (spatial) & Decentr. & Supervised + Attention & \VTwoX & \ConeTwo \\
		CoBEVT~\cite{xu2023cobevt} & \ygray{2023} & \InterBadge & Broadcast (Transf.) & Centr. & Supervised & \VTwoX & \Cone \\
		\rowcolor{gray!6} Fusion2comm~\cite{chu2025occlusion} & \ygray{2025} & \InterBadge & Saliency (occlusion) & Decentr. & Supervised + Attention & \VTwoX & \ConeTwo \\
		Which2comm~\cite{yu2025which2comm} & \ygray{2025} & \InterBadge & Saliency (multi-scale) & Decentr. & Supervised + Attention & \VTwoX & \ConeTwo \\
		\rowcolor{gray!6} QuantV2X~\cite{zhao2025quantv2x} & \ygray{2025} & \InterBadge & Quantization & Decentr. & Supervised + Quant. & \VTwoX & \ConeTwo \\
		CoGMP~\cite{fu2025generative} & \ygray{2025} & \InterBadge & Gen. Reconstr. & Centr. & Diffusion & \VTwoX & \ConeTwo \\
		\rowcolor{gray!6} V2XPnP~\cite{zhou2025v2xpnp} & \ygray{2025} & \InterBadge & Spatiotemporal & Centr. & Supervised & \VTwoX & \Cone \\
		InfoCom~\cite{wei2026infocom} & \ygray{2026} & \InterBadge & IB-driven pruning & Decentr. & Supervised + IB & \VTwoX & \ConeTwo \\
		\midrule
		\multicolumn{8}{l}{\textit{Negative exemplars (failure regimes from Section~\ref{sec:open}, shown to make the criteria visibly discriminating):}}\\
		Latency-saturated early fusion & \ygray{--} & \EarlyBadge & Broadcast (stale) & Centr. & -- & \VTwoX & \Cnone \\
		\rowcolor{gray!6} Byzantine-corrupted fusion & \ygray{--} & \InterBadge & Corrupted & Decentr. & -- & \VTwoX & \Cnone \\
		\bottomrule
	\end{tabularx}
	\smallskip
	\parbox{\linewidth}{\footnotesize \textit{Abbreviations:} \InterBadge\ = Intermediate, Decentr. = Decentralized, Centr. = Centralized, Gen. = Generative, Reconstr. = Reconstruction, Distill. = Distillation, Quant. = Quantization, IB = Information Bottleneck, GNN = Graph Neural Network, Transf. = Transformer. \textit{Synergy Conditions:} \Cone\ = Semantic Disambiguation, \Ctwo\ = Pragmatic Exchange, \Cthree\ = Proactive Foraging, \Cnone\ = satisfies none (anti-synergy failure regime).}
\end{threeparttable}
\end{table}


This section examines two orthogonal design dimensions of collaborative architectures that must be decoupled in analysis. The first is the \textbf{collaboration stage} (Section~\ref{sec:stage}), the abstraction level at which information is exchanged, which governs \emph{what} is shared. The second is the \textbf{fusion topology} (Section~\ref{sec:topology}), the physical and logical structure through which agents route and aggregate shared information, which governs \emph{to whom} it is sent. 
The logical frontier of this paradigm lies in software-defined collaborative networks (SDCN) (Section~\ref{sec:sdcn}), which adapt \emph{both} dimensions dynamically in response to real-time mission constraints, wireless propagation environments, and computational loads. Table~\ref{tab:taxonomy_arch} provides a consolidated comparison of representative methods spanning these architectural dimensions. This section covers Dimensions~1 and~3 of the taxonomy of Section~\ref{sec:foundations}, advancing \Cone\ (Semantic Disambiguation) and \Ctwo\ (Pragmatic Exchange).

\subsection{Collaboration Stages}\label{sec:stage}

The collaboration stage determines the abstraction level at which information is exchanged among agents, directly governing the rate--fidelity trade-off formalized in Eq.~\ref{eq:arch_ib}. Three canonical stages span the informational spectrum.

\begin{figure}[t]
\begin{center}
	\includegraphics[width=\linewidth]{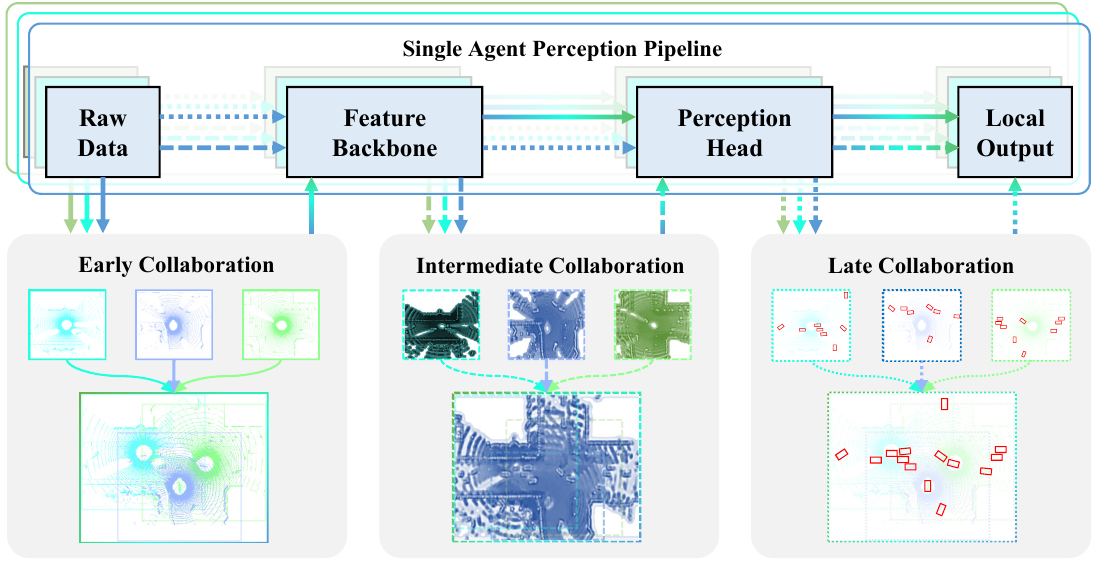}
\end{center}
\caption{\textbf{Illustration of three primary collaboration strategies in multi-agent perception systems.} The collaboration stage dictates the abstraction level at which information is exchanged. \emph{Early Collaboration} (solid line) shares high-bandwidth raw data before the feature backbone. \emph{Intermediate Collaboration} (dashed line) shares compressed features after the backbone, offering a balance between performance and communication cost. \emph{Late Collaboration} (dotted line) shares low-bandwidth local outputs (e.g., object detections) after the perception head. Lower panels illustrate the fusion process within each strategy, where data from multiple agents (differentiated by color) is aggregated into a unified representation before being re-injected into the ego-agent's pipeline.}
\label{fig:2_3}
\end{figure}

Traditional collaboration paradigms represent the asymptotic boundaries of the collaborative spectrum, as illustrated in Figure~\ref{fig:2_3}. While they provide foundational benchmarks, they encounter fundamental scaling limitations in large-scale, dynamic multi-agent ecosystems.

\subsubsection{Early collaboration}

Early collaboration refers to raw data aggregation that functions as the upper informational bound of the system. In this paradigm, agents act as transparent conduits, transmitting uncompressed 3D point clouds or high-definition RAW video streams to a central aggregator. By preserving the full raw entropy of the environment, early collaboration allows for global joint optimization without the quantization loss inherent in local feature extraction, enabling the aggregator to perform fine-grained spatial reasoning when the communication link can sustain the raw data load~\cite{chen2019cooper}. 
Despite its theoretical appeal, however, early collaboration is often impractical for real-time mobile swarms. The communication load creates a latency penalty. In high-velocity scenarios such as highway platooning or agile aerial inspection, this induces temporal registration error: by the time the central node fuses the raw data, the physical state of the world may have already evolved, rendering the fused manifold a delayed estimate rather than a reliable guide for immediate action. Consequently, early collaboration is mainly suitable for short-range high-bandwidth links, tethered multi-robot systems, or offline mapping settings.

\subsubsection{Late collaboration}

In contrast to early collaboration, late collaboration represents the limit of informational abstraction. Agents exchange only high-level metadata, such as 3D bounding boxes, semantic class labels, and localized trajectory coefficients~\cite{yu2022dair}. The primary advantage of late collaboration is its hardware-agnostic nature: a LiDAR-equipped rover can easily collaborate with a low-cost monocular drone by simply comparing object lists, creating a highly scalable and resilient network that is immune to raw data heterogeneity. However, late collaboration usually suffers from the hard thresholding error. Because collaboration occurs after the local agent has made a final decision, it lacks the evidential depth to resolve the local observation trap. If an agent's local detector fails due to severe sensor noise or adversarial occlusion, it broadcasts a signal with detection errors, and the fusion center has no access to the low-confidence feature signals that may allow a neighbor to correct the error.

\subsubsection{Intermediate collaboration}

Intermediate collaboration redefines collaboration as task-oriented latent feature exchange, occupying the Pareto-optimal region of the rate--fidelity frontier formalized in Eq.~\ref{eq:arch_ib}. Rather than exchanging explicit geometric primitives or sparse metadata, agents share compressed neural embeddings that preserve task-relevant semantics while discarding sensor-specific redundancy~\cite{chen2019f, wang2020v2vnet, xu2022v2x, zhang2024collaborative}. Complementary strategies include region-based hybrid fusion that partitions the sensing field by spatial saliency~\cite{liu2023region}, collective perception schemes that treat the swarm as a distributed sensor array~\cite{pilz2023collective}, and adaptive architectures that dynamically adjust collaboration depth in response to channel quality and task criticality.

Advanced frameworks such as V2X-ViT~\cite{xu2022v2x} and CoBEVT~\cite{xu2023cobevt} utilize cross-attention in Transformers to project disparate, multi-perspective features into a unified bird's-eye-view (BEV) manifold. This architecture facilitates a generative process known as collaborative reconstruction~\cite{wang2023core}, in which high-confidence features from one viewpoint are used to fill the semantic gaps of an occluded neighbor. Mathematically, this is equivalent to a probabilistic prior injection, where the collective knowledge of the swarm reconstructs the obscured features of an actor, effectively turning the swarm into a distributed panoramic sensor.

The next frontier of intermediate collaboration involves the transition from periodic broadcasting to pragmatic informational foraging. The foundational step along this path was to ask \emph{who} and \emph{when} to communicate: Who2com~\cite{liu2020who2com} selects collaborators through a learnable handshake protocol, while When2com~\cite{liu2020when2com} groups the communication graph and activates message exchange only when it benefits the collective, recasting communication itself as a selective, learnable decision rather than a fixed broadcast. Frameworks such as Where2comm~\cite{hu2022where2comm} and Which2comm~\cite{yu2025which2comm} treat the communication channel as a strategic resource, utilizing spatial confidence maps and epistemic uncertainty markers to identify feature patches that provide the maximum reduction in global world-model entropy. How2comm~\cite{yang2023how2comm} extends this line with mutual information-driven spatial-channel message filtering and flow-guided temporal delay compensation, addressing communication redundancy, transmission latency, and heterogeneous agent features within a unified spatiotemporal collaborative Transformer. At the extreme of compression, InfoCom~\cite{wei2026infocom} reformulates collaborative perception through an extended information bottleneck (IB) framework, employing information-aware encoding, sparse mask generation with 4-bit quantization, and multi-scale BEV decoding to achieve near-lossless perception at kilobyte-scale transmission, a 440$\times$ reduction over Where2comm and over 4,000$\times$ reduction compared with standard collaborative perception baselines, as reported in~\cite{wei2026infocom}. By transmitting only high-entropy or mission-critical regions (e.g., an occluded pedestrian at a blind intersection), these architectures achieve extreme data reduction while maintaining the structural integrity of the collective model~\cite{hu2026pragmatic}. This pragmatic communication strategy ensures the swarm transmits only decision-relevant information, maximizing the utility of each bit under bandwidth constraints.

Generative map priors (GMP) represent a paradigm shift from discriminative detection to generative reconstruction in the collaborative manifold. As autonomous swarms operate in high-clutter environments, traditional fusion often fails when critical perspectives are missing. To bridge this gap, CI architectures leverage latent diffusion models (LDMs) to synthesize a high-fidelity, globally consistent world model by conditioning on sparse, distributed cues.

Unlike standard pixel-level diffusion, collaborative LDMs operate within the compressed neural manifold to ensure real-time feasibility. Frameworks such as CoGMP~\cite{fu2025generative} utilize a multi-agent diffusion prior to denoise the incomplete BEV features. When a swarm experiences a significant loss of perspective, the generative decoder treats the remaining latent features as sparse anchors. Through a forward diffusion process that injects Gaussian noise and a subsequent reverse denoising process, the model reconstructs the missing semantic geometry. The iterative refinement ensures that reconstructed regions are not merely random textures, but are topologically consistent with the observed spatial constraints of the collective.

The generative capability of CI is further enhanced by integrating spatiotemporal memory banks. By conditioning the diffusion process on historical priors, the swarm can maintain the identity persistence of an actor when it is occluded from all other perspectives, enabling the swarm to reconstruct the probable trajectory and semantic state of hidden objects based on their last known vectors and the collective's shared understanding of environmental physics. To guard against the over-generation risk inherent in generative world models, reconstructed regions are paired with evidential uncertainty estimates so that high-uncertainty voxels are flagged for active verification, closing the loop between generation and sensing. Complementary evidential deep learning approaches assign Dirichlet distributions over shared voxels to parameterize this uncertainty without Monte Carlo sampling~\cite{yuan2023generating}.

Generative map priors also reconcile heterogeneous sensor inputs. For instance, when a drone's RGB camera is blinded by glare, the generative world model utilizes the sparse LiDAR pulses from a ground rover to synthesize the missing visual semantics. This cross-modal generative synthesis ensures that the global manifold remains informationally saturated even under extreme sensor-specific degradation, effectively creating a weather-invariant and modality-agnostic collective sight~\cite{li2025v2x, kou2025adverse}. It should be noted, however, that the benefits of intermediate collaboration are not unconditional: when communication cost, generative over-generation risk, or correlated sensor failures dominate, collaborative perception can degrade below single-agent baselines; these failure modes are analyzed in Section~\ref{sec:open}.

\subsection{Architecture Topologies}\label{sec:topology}

\begin{figure}[t]
	\begin{center}
		\includegraphics[width=\linewidth]{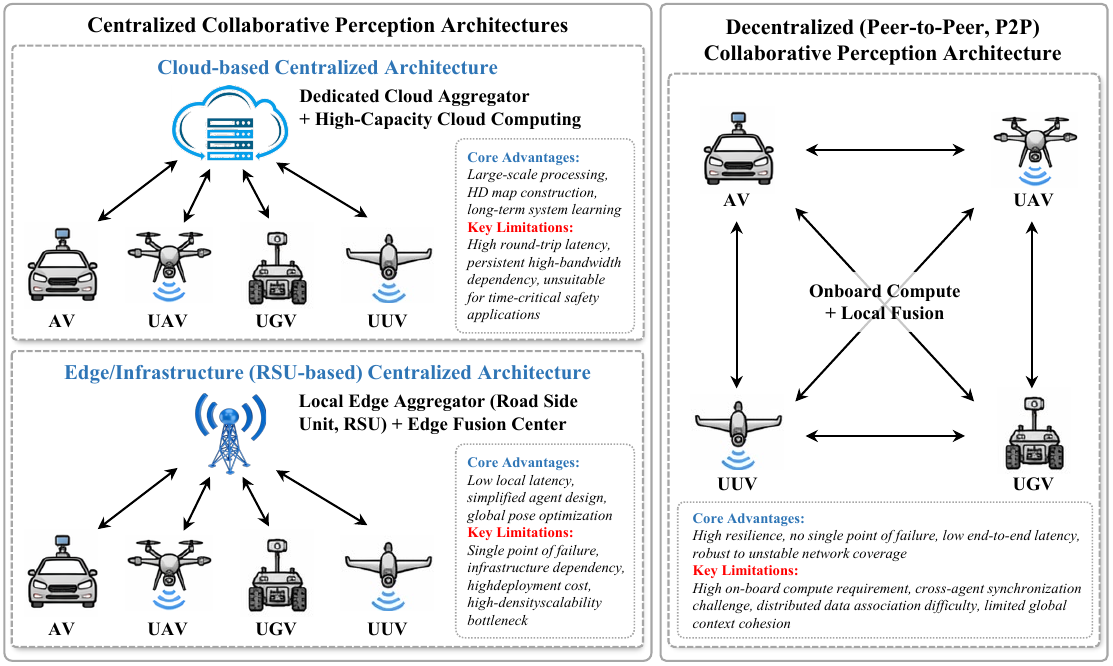}
	\end{center}
	\caption{\textbf{Canonical collaborative perception architecture topologies.} (a) Cloud-based centralized: remote data center aggregation, global coverage at high latency. (b) Edge/infrastructure-based: RSU local fusion, low latency with infrastructure dependency. (c) Decentralized P2P: on-board V2V fusion, high resilience at the cost of compute and synchronization overhead. The SDCN paradigm (Section~\ref{sec:sdcn}) targets adaptive transitions among these topologies.}
	\label{fig:arch}
\end{figure}

Collaboration stages determine at what abstraction level information is exchanged; a second, orthogonal design dimension is the physical topology through which that information is routed and aggregated. These topologies can in principle be combined with any collaboration stage; for instance, both early and intermediate fusion can be realized in centralized, edge-based, or decentralized configurations~\cite{bai2024survey}. Figure~\ref{fig:arch} contrasts the three canonical topologies.

\noindent\textbf{Cloud-based centralized.} In the cloud-based topology, a remote data center aggregates raw or processed sensor streams from all agents, enabling global-scale processing, HD map construction, and long-term learning across large fleets~\cite{gao2024vehicle}. The fundamental limitation is round-trip latency, which typically amounts to hundreds of milliseconds over cellular backhaul and renders this topology unsuitable for time-critical safety applications such as collision avoidance. It is therefore primarily deployed in offline mapping, fleet analytics, and training-data curation pipelines rather than in real-time control loops.

\noindent\textbf{Edge/infrastructure-based.} The edge-based topology deploys roadside units (RSUs) as local fusion centers, striking a balance between global coordination and local responsiveness. RSU-based architectures offer low local latency, simplified agent-side design (offloading fusion computation to infrastructure), and globally optimized pose alignment through fixed, calibrated sensor mounts~\cite{zhang2023roadside,hao2024rcooper}. However, they introduce single-point-of-failure vulnerability at each RSU, high deployment and maintenance cost, and scalability bottlenecks as the number of agents within an RSU's coverage area grows. The tension between coverage density and infrastructure cost remains an open challenge.

\noindent\textbf{Decentralized peer-to-peer.} In the decentralized P2P topology, each agent performs fusion on-board using direct V2V or agent-to-agent links. This architecture achieves high resilience with no single point of failure, minimal end-to-end latency (no infrastructure relay), and inherent scalability through local neighborhood communication~\cite{wang2020v2vnet}. The trade-offs include substantial on-board compute requirements, challenging cross-agent synchronization without a central clock reference, and limited global context cohesion when the communication graph is sparse~\cite{han2023collaborative}. Decentralized topologies are particularly well-suited for high-mobility scenarios where infrastructure coverage is intermittent or unavailable.

\subsection{Software-Defined Collaborative Networks (SDCN)}\label{sec:sdcn}

A promising direction for CI architectures is the realization of the software-defined collaborative network (SDCN). In this paradigm, the structural organization of GCI is no longer a static choice but an adaptive meta-cognitive state~\cite{xu2025cosdh,bai2024survey}.

\noindent\textbf{Cyber-Physical Context Awareness.} In an SDCN, the parameters of collaboration, i.e., fusion depth, compression ratio, and feature saliency, are dynamically modulated by a meta-cognitive layer. This layer continuously monitors the cyber-physical health of the swarm, including the network state (jitter, bandwidth availability, and packet loss rates), the computational thermal limits (onboard energy reserves and FLOP availability), and the mission criticality (proximity to safety-critical hazards or the importance of the current objective)~\cite{xue2025joint,xiao2023perception}.

\noindent\textbf{Adaptive Fusion Elasticity.} When the swarm operates in a low-risk, bandwidth-constrained environment (e.g., open-sea mapping), the SDCN autonomously shifts toward hyper-sparse late collaboration to conserve energy. Conversely, when the swarm enters a high-risk zone (e.g., a congested, unsignalized urban intersection), the system undergoes phase transition, shifting toward dense neural manifold synthesis. This elastic collaboration ensures that the swarm maintains a robust global semantic manifold in the most contested and resource-constrained physical environments~\cite{xu2025cosdh,liu2025mmcooper}.

\noindent\textbf{Self-Organizing Perceptual Topologies.} In an SDCN, the communication graph can be dynamically reconfigured, with perceptual hubs elected based on vantage point quality and available computational headroom. This represents the convergence of sensing, communication, and reasoning into a unified architectural framework that improves resilience and adaptivity under changing network and mission conditions~\cite{zhang2025bridging}.



\section{Neural-Communication Co-Design}\label{sec:comm}

\begin{table}[t]
	\centering
	\begin{threeparttable}
		\caption{\textbf{Taxonomy of neural-communication co-design methods and cognitive synergy conditions.}}
		\label{tab:taxonomy_comm}
		\renewcommand{\arraystretch}{1.5}
		\tiny
		\setlength{\tabcolsep}{4pt}
		\renewcommand{\tabularxcolumn}[1]{m{#1}}
		\sffamily
		\begin{tabularx}{\textwidth}{
				>{\raggedright\arraybackslash\hsize=1.3\hsize}X
				>{\centering\arraybackslash\hsize=0.7\hsize}X
				>{\centering\arraybackslash\hsize=0.8\hsize}X
				>{\raggedright\arraybackslash\hsize=1.5\hsize}X
				>{\centering\arraybackslash\hsize=0.8\hsize}X
				>{\raggedright\arraybackslash\hsize=1.5\hsize}X
				>{\centering\arraybackslash\hsize=0.7\hsize}X
				>{\centering\arraybackslash\hsize=0.7\hsize}X}
			\toprule
			\textbf{Method} & \textbf{Year} & \textbf{Stage} & \textbf{Comm.\ Paradigm} & \textbf{Architecture} & \textbf{Learning Strategy} & \textbf{Application} & \textbf{Synergy} \\
			\midrule
			Deep JSCC~\cite{bourtsoulatze2019deep, yao20256g} & \ygray{2019} & \InterBadge & JSCC (SNR-adaptive) & Any & Supervised + JSCC & \General & \Ctwo \\
			\rowcolor{gray!6} NTSCC~\cite{dai2022nonlinear} & \ygray{2022} & \InterBadge & JSCC (nonlinear transform) & Any & Supervised + JSCC & \General & \Ctwo \\
			CoBEVFlow~\cite{wei2023asynchrony} & \ygray{2023} & \InterBadge & Flow-based temporal comp. & Centr. & SSL + Optical Flow & \VTwoX & \ConeTwo \\
			\rowcolor{gray!6} HADCoP~\cite{thornton2025real} & \ygray{2025} & \InterBadge & On-demand (DRL) & Decentr. & MARL + JSCC & \VTwoX & \ConeTwo \\
			PragComm~\cite{hu2026pragmatic} & \ygray{2026} & \InterBadge & IB-driven (Lagrangian) & Any & IB-optimized & \MultiDom & \ConeTwo \\
			\bottomrule
		\end{tabularx}
		\smallskip
		\parbox{\linewidth}{\footnotesize \textit{Abbreviations:} JSCC = Joint Source-Channel Coding, IB = Information Bottleneck, DRL = Deep Reinforcement Learning, SSL = Self-Supervised Learning, Centr.\ = Centralized, Decentr.\ = Decentralized, Gen.\ = general-purpose method not tied to a specific application domain. \textit{Synergy Conditions:} \Cone\ = Semantic Disambiguation, \Ctwo\ = Pragmatic Exchange, \Cthree\ = Proactive Foraging.}
	\end{threeparttable}
\end{table}

\begin{figure}[t]
	\begin{center}
		\includegraphics[width=\linewidth]{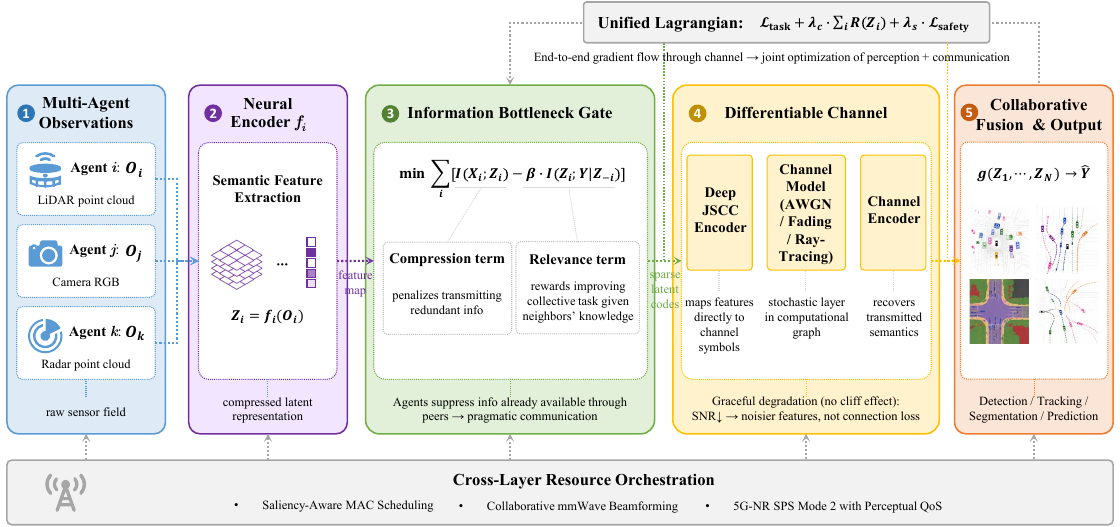}
	\end{center}
	\caption{\textbf{Neural-communication co-design pipeline for collaborative perception.} The pipeline transforms raw multi-agent observations into a collaborative perception output through five stages. (1) Multi-modal sensor observations from heterogeneous agents (LiDAR, camera, radar). (2) Neural feature encoding that compresses raw streams into compact latent representations. (3) An information bottleneck (IB) gate that enforces pragmatic communication by jointly minimizing transmitted redundancy while maximizing task-relevant mutual information given neighbors' knowledge (Eq.~\ref{eq:ib_collaborative}). (4) A differentiable channel model (Deep JSCC) that directly maps latent codes to channel-modulated symbols, enabling end-to-end gradient flow from the task loss back through the physical layer and thereby avoiding the cliff effect. (5) Collaborative fusion that produces the multi-agent BEV detection, tracking, segmentation, and prediction output. The top feedback loop encodes the unified joint perception--communication Lagrangian (Eq.~\ref{eq:unified_lagrangian}), enabling the entire stack to be optimized as a single differentiable system.}
	\label{fig:neur_comm}
\end{figure}

The classical separation theorem of source and channel coding treats sensing, compression, and transmission as independent blocks. GCI in dynamic environments demands departure from this decoupled approach toward neural-communication co-design, which treats the channel as a differentiable pipeline component, ensuring bandwidth serves the features that maximize collective task performance~\cite{yao20256g, hu2026pragmatic}. Neural-communication co-design spans a spectrum from information-theoretic encoding to real-time resource allocation at the physical layer, as illustrated in Figure~\ref{fig:neur_comm}. This section covers Dimension~2 (Communication Paradigm) and realizes \Ctwo\ (Pragmatic Exchange).

\subsection{Deep Joint Source-Channel Coding}

Traditional V2X frameworks rely on a compress-then-transmit strategy, which is notoriously brittle under low signal-to-noise ratio (SNR) or high-interference conditions; recent 6G-era cooperative perception work~\cite{yao20256g, research0342} re-examines this pipeline under tighter latency and fidelity targets. Deep joint source-channel coding (Deep JSCC) reimagines this by mapping high-dimensional sensory data directly into channel-modulated symbols, enabling the system to learn an optimal mapping from the source distribution to the channel input space without explicit quantization or entropy coding.

\noindent\textbf{Information Bottleneck Formulation for Collaborative Perception.} The theoretical foundation of Deep JSCC in collaborative settings can be formalized through the information bottleneck (IB) principle~\cite{hu2026pragmatic}. Using the shared notation of Section~\ref{sec:foundations}, let \(\mathbf{O}_i\) denote the raw sensory observation of agent \(i\), \(\mathbf{Z}_i\) the transmitted latent representation, and \(\mathbf{Y}\) the global world state (or the downstream collaborative task derived from it, e.g., 3D object detection or semantic segmentation). The IB objective for the collective is:
\begin{equation}
\min \sum_{i} \Big[ I(\mathbf{O}_i; \mathbf{Z}_i) - \beta \, I(\mathbf{Z}_i; \mathbf{Y} \mid \mathcal{Z}_{-i}) \Big],
\label{eq:ib_collaborative}
\end{equation}
where \(I(\cdot;\cdot)\) denotes mutual information, \(\mathcal{Z}_{-i}\) is the set of latents received from other agents, and \(\beta\) controls the rate--distortion trade-off. The first term penalizes transmitting redundant or task-irrelevant information (compression), while the second term rewards latents that improve the collective's task performance \emph{given what neighbors already know} (relevance). This conditional formulation is the key distinction from single-agent IB: an agent is encouraged to suppress information already available through its peers, naturally inducing a pragmatic communication policy where only \emph{informational surprise} is transmitted. In practice, the conditional mutual information \(I(\mathbf{Z}_i; \mathbf{Y} \mid \mathcal{Z}_{-i})\) is intractable to compute exactly in high-dimensional feature spaces. Feasible approximations include \emph{variational IB}, which replaces the intractable conditional distribution with a learned variational surrogate (e.g., a neural estimator of the mutual information gap), and \emph{contrastive estimation}, where the relevance term is approximated by discriminating positive feature pairs (from agents observing the same scene) from negative pairs (from different scenes). These approximations follow standard practice in deep IB and mutual-information neural estimation and are discussed further in the training methodology subsection below.

\noindent\textbf{SNR-Adaptive Neural Encoders.} Unlike static compression algorithms (e.g., H.264 or JPEG), co-designed neural encoders utilize attention mechanisms to perform semantic power allocation. Architectures inspired by Deep JSCC~\cite{yao20256g} and nonlinear transform source-channel coding (NTSCC)~\cite{dai2022nonlinear} replace the traditional separate-source-channel-coding pipeline with a single, end-to-end optimized neural codec. When an agent detects a high-priority object, such as a pedestrian in a blind spot, the encoder dynamically scales the energy of the corresponding latent features. This ensures that even if the physical channel experiences deep fading, the critical semantic tokens remain decodable by the fusion center. This graceful degradation, known in the JSCC literature as \emph{cliff effect} avoidance, allows the swarm to maintain situational awareness in weather conditions or urban canyons where traditional digital protocols would suffer from catastrophic packet loss~\cite{bourtsoulatze2019deep, dai2022nonlinear}.

\noindent\textbf{Task-Oriented Feature Pruning.} In a co-designed network, the objective is not pixel-level reconstruction but downstream task performance. By integrating the IB principle (Eq.~\ref{eq:ib_collaborative}) directly into the training objective, the neural encoder learns to suppress background entropy that is irrelevant to the collective mission. Concretely, the encoder produces an extremely sparse feature manifold where only information that is both \emph{task-relevant} and \emph{not already known} to neighbors is transmitted. This pragmatic communication reduces the total bandwidth requirement by orders of magnitude compared to raw data sharing while maintaining near-identical perception accuracy~\cite{hu2022where2comm, hu2026pragmatic}. Saliency-guided transmission achieves over 90\% communication volume reduction at near-identical accuracy~\cite{hu2022where2comm, yu2025which2comm}, and end-to-end quantization further reduces latency by 3.2$\times$ while improving mAP by 9.5 over full-precision baselines~\cite{zhao2025quantv2x}. 

\subsection{Training Methodologies}

Training a neural-communication co-designed system presents unique challenges beyond standard deep learning, stemming from the need to jointly optimize perception and communication objectives across a distributed architecture with non-differentiable channel components.

\noindent\textbf{End-to-End Learning with Differentiable Channel Models.} The core enabler of co-design is the differentiable channel model. During training, the physical channel (including modulation, fading, and noise) is represented as a stochastic layer within the computational graph. This allows gradients from the task loss (e.g., detection or segmentation error) to flow backward through the channel, jointly optimizing the neural encoder, power allocation, and downstream fusion modules. In practice, channel models range from simple additive white Gaussian noise (AWGN) approximations for initial prototyping~\cite{bourtsoulatze2019deep} to ray-tracing-based digital twins that capture environment-specific multipath and blockage effects for deployment-grade training~\cite{cazzella2024multi}.

\noindent\textbf{Alternating Optimization and Surrogate Gradients.} When the channel is non-differentiable (e.g., in real-world over-the-air training or when using hardware-specific modulation), two strategies are commonly employed. \emph{Alternating optimization} decouples the problem: the perception stack is trained with an idealized channel, then the communication policy (e.g., power allocation, scheduling) is optimized via reinforcement learning using the perception model as a fixed reward function~\cite{ghnaya2023multi}. \emph{Surrogate gradient methods} replace the non-differentiable channel operations with smooth approximations during the backward pass~\cite{tung2022deepjscc}, enabling approximate end-to-end training without requiring an exact channel Jacobian. The choice between these strategies depends on the degree of channel variability: surrogate gradients excel in slowly fading channels, while alternating optimization is more robust under rapid fluctuation.

\noindent\textbf{Multi-Agent Credit Assignment in Communication.} A further challenge is assigning credit for communication decisions: \emph{if the collective detection improves, which agent's transmission is responsible?} Recent works address this via counterfactual communication gradients~\cite{su2020counterfactual}, where each agent's contribution to the task loss is estimated by comparing the actual fused output against a counterfactual baseline~\cite{foerster2018counterfactual} that excludes that agent's transmission. This provides a per-agent training signal that encourages agents to transmit only when their observation carries unique, decision-critical information.

\subsection{Channel-Aware Adaptive Fusion}

The neural-communication synergy extends to the fusion architecture, which tends to be elastic enough to adapt to fluctuating network topologies and latencies.

\noindent\textbf{Dynamic Model Partitioning and Offloading.}
GCI architectures aim to resolve the tension between onboard energy constraints and the need for high-fidelity global fusion. Co-designed systems utilize adaptive partitioning, where the cutoff point between local computation and shared processing is shifted in real-time~\cite{thornton2025real}. Under high-throughput conditions, agents offload low-level feature extraction to an edge server or roadside unit to leverage multi-perspective Transformer blocks. During network congestion, the agents autonomously shift toward late collaboration, performing more heavy-duty processing locally to minimize the informational payload. This cyber-physical elasticity ensures that CI remains functional when the communication fabric is partially compromised~\cite{xu2025cosdh}.

\noindent\textbf{Heterogeneous Feature Alignment Under Jitter.}
Real-world communication is plagued by asynchronous packet arrivals and clock jitter. A co-designed system does not wait for all packets to arrive, which can induce unacceptable latency. Instead, it utilizes neural asynchrony compensation. The fusion block treats incoming feature tensors as spatiotemporal samples and uses learned motion cues to warp delayed features into the current timeframe~\cite{wei2023asynchrony}. This neural asynchrony compensation ensures that the global world model remains temporally consistent despite the underlying network's jitter or retransmission delays~\cite{lei2022latency}.

\subsection{Cross-Layer Resource Orchestration}

Another dimension of co-design involves the direct control of physical (PHY) and media access control (MAC) layer resources by the perceptual layer.

\noindent\textbf{Saliency-Aware Medium Access.} Moving beyond data-agnostic MAC protocols, co-designed GCI links channel access to perceptual surprise: high-risk or novel detections receive elevated scheduling priority so safety-critical information is transmitted before routine updates, bridging the gap between existing 5G-NR QoS frameworks and perception-aware resource allocation.

\noindent\textbf{Collaborative Beamforming for Semantic Feature Exchange.}
When directional links are available, unmanned aerial vehicle (UAV) or vehicular collectives can align communication resources with sensing value. Agents that observe high-entropy zones (e.g., a complex intersection or disaster site) can be allocated stronger beams, more reliable links, or higher scheduling priority. In mmWave and sub-THz bands, where beam management is already a core 5G-NR function (SSB-based beam sweeping and CSI-RS beam refinement), the perception module can directly inform beam selection: rather than performing exhaustive spatial sweeping, the agent uses its onboard perception output to steer the communication beam toward the spatial sector where collaborating agents are expected to benefit most from the shared feature payload. This approximates a perception-aware distributed MIMO configuration in which physical and communication topology are optimized together~\cite{guo2024predictive}.

\subsection{Synthesis}

While the co-design paradigm offers substantial gains, its practical deployment is bounded by the realism of the channel models and the stability of multi-agent training, as discussed in the critical reflection on sim-to-real limitations in Section~\ref{sec:open}. Across the integration levels surveyed above, a common mathematical theme emerges: the co-design optimization objective can be expressed as a \emph{joint perception--communication Lagrangian}:
\begin{equation}
\mathcal{L} = \underbrace{\mathcal{L}_{\text{task}}(\mathbf{Y}, \hat{\mathbf{Y}})}_{\text{Perception fidelity}} \;+\; \lambda_c \underbrace{\sum_{i} R(\mathbf{Z}_i)}_{\text{Communication cost}} \;+\; \lambda_s \underbrace{\mathcal{L}_{\text{safety}}(\hat{\mathbf{Y}})}_{\text{Safety constraint}},
\label{eq:unified_lagrangian}
\end{equation}
where \(\mathcal{L}_{\text{task}}\) is the downstream task loss, \(R(\mathbf{Z}_i)\) the transmission rate of agent \(i\)'s latent code, and \(\mathcal{L}_{\text{safety}}\) a penalty encoding safety-critical constraints (e.g., minimum detection recall for vulnerable road users). The multipliers \(\lambda_c\) and \(\lambda_s\) are not fixed hyperparameters but are dynamically modulated by the meta-cognitive layer of a software-defined collaborative network (Section~\ref{sec:sdcn}), adapting to real-time cyber-physical context.

This unified calculus reveals three critical insights for scaling GCI systems. First, the perception--communication boundary is a \emph{design choice}, not a given. The optimal partition point between local computation and shared transmission shifts continuously with channel conditions and mission criticality. Second, the IB formulation (Eq.~\ref{eq:ib_collaborative}) provides a principled mechanism for pragmatic communication: agents learn to transmit only what their neighbors do not already know, naturally implementing the pragmatic information-exchange criterion in Table~\ref{tab:synergy_operational}. Third, the integration of safety constraints (\(\mathcal{L}_{\text{safety}}\)) directly into the co-design objective ensures that bandwidth allocation under congestion is \emph{safety-aware} rather than merely throughput-optimal, a distinction that is critical for the trustworthy synergy discussed in Section~\ref{sec:trust}.

Key open challenges remain. The non-stationarity of multi-agent wireless channels, in which each agent's transmission is interference to the others, introduces a complex co-adaptation problem that can lead to training instability. Scaling differentiable channel models to capture the full richness of mmWave or THz propagation (including beam management and blockage prediction) remains an active research frontier. Finally, interoperable semantic payload descriptions are a prerequisite for plug-and-play feature exchange across heterogeneous fleets. Resolving these challenges will transform the communication fabric from an external constraint into an integral, optimized dimension of the collaborative intelligence stack. From an engineering-realizability standpoint, the co-design Lagrangian of Eq.~\ref{eq:unified_lagrangian} is tractable only under explicit resource budgets: V2X field deployments typically constrain end-to-end perception latency to 30--50~ms and per-link bandwidth to tens of KB/frame, which forces $\lambda_c$ toward regimes where most candidate features are suppressed and only mission-critical patches survive the IB gate. Foundation-model-based collaboration sharpens this tension rather than relaxing it: autoregressive decoding and multi-round large language model (LLM) debate add latency orders of magnitude above the control loop budget, so practical systems must cache pretrained representations, quantize the shared backbone, and confine language-level reasoning to non-time-critical layers (e.g., mission replanning) while the perception-communication stack continues to operate at sub-second cadence. These deployable operating points, not the theoretical optimum of the Lagrangian, define the realistic envelope within which GCI can be fielded today.

\section{Embodied Synergy}\label{sec:embodied}

The methods surveyed in this section span collaborative active perception, multi-agent reinforcement learning for actuation synergy, and shared intent and predictive coordination, as organized in Table~\ref{tab:taxonomy_embodied}.

\begin{table}[t]
	\centering
	\begin{threeparttable}
		\caption{\textbf{Taxonomy of embodied synergy methods and cognitive synergy conditions.}}
		\label{tab:taxonomy_embodied}
		\renewcommand{\arraystretch}{1.5}
		\tiny
		\setlength{\tabcolsep}{1pt}
		\renewcommand{\tabularxcolumn}[1]{m{#1}}
		\sffamily
		\begin{tabularx}{\textwidth}{
				>{\raggedright\arraybackslash\hsize=1.3\hsize}X
				>{\centering\arraybackslash\hsize=0.7\hsize}X
				>{\centering\arraybackslash\hsize=0.8\hsize}X
				>{\raggedright\arraybackslash\hsize=1.4\hsize}X
				>{\centering\arraybackslash\hsize=0.9\hsize}X
				>{\raggedright\arraybackslash\hsize=1.4\hsize}X
				>{\centering\arraybackslash\hsize=0.7\hsize}X
				>{\centering\arraybackslash\hsize=0.7\hsize}X}
			\toprule
			\textbf{Method} & \textbf{Year} & \textbf{Stage} & \textbf{Comm.\ Paradigm} & \textbf{Architecture} & \textbf{Learning Strategy} & \textbf{Application} & \textbf{Synergy} \\
			\midrule
			CooperNaut~\cite{cui2022coopernaut} & \ygray{2022} & \InterBadge & On-demand & Decentr. & MARL & \VTwoX & \ConeTwoThree \\
			\rowcolor{gray!6} JFP~\cite{luo2023jfp} & \ygray{2023} & \InterBadge & Broadcast & Decentr. & MARL & \VTwoX & \ConeTwoThree \\
			Plan2comm~\cite{xie2024towards} & \ygray{2024} & \InterBadge & Saliency + Planning & Decentr. & MARL + Supervised & \VTwoX & \ConeTwoThree \\
			\rowcolor{gray!6} CoPnP~\cite{ren2024collaborative} & \ygray{2024} & \InterBadge & On-demand & Decentr. & MARL & \VTwoX & \ConeTwoThree \\
			CMP~\cite{wang2025cmp} & \ygray{2025} & \InterBadge & Saliency & Decentr. & MARL + Supervised & \VTwoX & \ConeTwoThree \\
			\rowcolor{gray!6} UniV2X~\cite{yu2025end} & \ygray{2025} & \InterBadge & On-demand & Hybrid & Supervised + MARL & \VTwoX & \ConeTwoThree \\
			CSAOT~\cite{nguyen2025csaot} & \ygray{2025} & \InterBadge & On-demand (VoI) & Decentr. & MARL + Attention & \UxS & \ConeTwoThree \\
			\rowcolor{gray!6} CooperDrive~\cite{qu2026cooperdrive} & \ygray{2026} & \InterBadge & On-demand & Decentr. & MARL & \VTwoX & \ConeTwoThree \\
			\bottomrule
		\end{tabularx}
		\smallskip
		\parbox{\linewidth}{\footnotesize \textit{Abbreviations:} VoI = Value of Information, MARL = Multi-Agent Reinforcement Learning, Decentr.\ = Decentralized. \textit{Synergy Conditions:} \Cone\ = Semantic Disambiguation, \Ctwo\ = Pragmatic Exchange, \Cthree\ = Proactive Foraging.}
	\end{threeparttable}
\end{table}

The evolution of GCI is currently navigating a fundamental transition: moving from the passive synthesis of a global semantic manifold to the execution of coordinated actuation in unconstrained physical environments. This progression necessitates a shift from a perception-only paradigm toward a tightly coupled action-perception loop. In this context, embodied synergy refers to the collective ability of a swarm to treat its physical movements and spatial configurations as strategic tools to actively reduce the entropy of its internal world model. By closing this loop, the swarm transforms from a group of distributed observers into an adaptive sensing-and-action system capable of purposeful environmental interaction. However, embodied coordination also introduces risks of cascading actuation errors and communication-induced instability, particularly when multi-agent reinforcement learning (MARL) policies trained in simulation are deployed without hardware-in-the-loop validation; these failure modes are examined in Section~\ref{sec:open}. This section examines the mechanisms that enable embodied synergy along three axes. We first formalize \textbf{collaborative active perception}, where the swarm treats its physical distribution as a controllable variable for information gain. We then examine \textbf{multi-agent reinforcement learning} architectures that map the shared neural manifold to distributed control signals. Finally, we discuss \textbf{shared intent prediction and heterogeneous coordination}, where agents jointly anticipate future states and exploit complementary physical niches.

\subsection{Collaborative Active Perception}

Traditional collaborative systems are predominantly reactive, processing whatever sensory data the environment happens to provide. However, in complex or adversarial scenarios, sensing and moving cannot be decoupled. This subsection operationalizes the third condition of cognitive synergy, i.e., \textbf{proactive informational foraging} (Section~\ref{sec:foundations}), by treating the swarm's physical distribution as a controllable variable to optimize information gain~\cite{dhami2024map}.

The formal objective of collaborative active perception is to select agent poses \(\{\mathbf{p}_i\}_{i=1}^{N}\) that maximize the expected information gain about the world state \(\mathbf{Y}\):
\begin{equation}
\max_{\mathbf{p}_1, \ldots, \mathbf{p}_N} \; I\bigl(\mathbf{Y}; \, \mathbf{O}_1(\mathbf{p}_1), \ldots, \mathbf{O}_N(\mathbf{p}_N)\bigr) \;-\; \sum_{i=1}^{N} C_{\text{move}}(\mathbf{p}_i, \mathbf{p}_i^{\text{curr}}),
\label{eq:active_perception}
\end{equation}
where \(\mathbf{O}_i(\mathbf{p}_i)\) is the observation that agent \(i\) collects at pose \(\mathbf{p}_i\), and \(C_{\text{move}}(\cdot)\) penalizes the movement cost from the current position \(\mathbf{p}_i^{\text{curr}}\). Equation~\ref{eq:active_perception} captures the dual nature of embodied synergy: the first term drives agents toward viewpoints that maximally resolve the collective's residual uncertainty (e.g., peering behind an occluding structure), while the second term ensures that repositioning is energy-efficient. In practice, online optimization requires learned approximations: MARL policies that map the current collective belief state to viewpoint-optimizing actions~\cite{westheider2023multi,selimovic2024multi}. For instance, in a disaster response scenario, an aerial drone may detect a high-uncertainty region behind a collapsed structure and autonomously command a ground rover to move to a specific coordinate that maximizes the perceptual dividend~\cite{yue2022aerial}.

Recent advances leverage MARL to train agents to act as active scouts. Frameworks like CSAOT~\cite{nguyen2025csaot} reward agents not for individual detection accuracy, but for their contribution to the collective's uncertainty reduction. This leads to the emergence of complex, coordinated behaviors such as tracking evasive targets or conducting top-down scouting, in which UAVs hover at specific altitudes to provide the optimal geometric prior for ground-based vehicles. This bidirectional flow, where perception guides movement, and movement optimizes perception, is the cornerstone of embodied resilience~\cite{suresh2024greedy}.

\subsection{Multi-Agent Reinforcement Learning}

Closing the loop requires mapping the shared neural manifold directly to distributed control signals. This is facilitated by MARL architectures that utilize the shared world representation (e.g., fused BEV features in V2X settings) as a common observation space for policy optimization~\cite{zhang2024multi}.

\noindent\textbf{Policy-Aware Feature Fusion and Intent Negotiation.}
Unlike standard fusion which aims for generic reconstruction, policy-aware fusion prioritizes features that directly influence the value function of the collective mission~\cite{yu2025end}. In high-stakes environments like unsignalized intersections, agents exchange intent-encoded latent features~\cite{parada2023end} to support decentralized negotiation of passing order and speed profiles; the domain-specific deployment of such intent sharing is discussed in Section~\ref{sec:domain}. Game-theoretic language is useful for formalizing these interactions, but practical systems still require explicit safety constraints, communication-delay handling, and fallback policies~\cite{pan2025cooperative, pei2021distributed}.

\noindent\textbf{End-to-End Cooperative Driving and Vision-Language Integration.}
A significant advance in closing the perception-action loop is the emergence of end-to-end cooperative driving frameworks that directly map shared sensory inputs to driving actions. Coopernaut~\cite{cui2022coopernaut} pioneers this direction by training an end-to-end model that fuses voxel-based LiDAR representations from multiple vehicles, demonstrating a 40\% improvement in navigation success rate over egocentric baselines on accident-prone scenarios while requiring five times less bandwidth than conventional late-fusion methods. More recently, V2X-VLM~\cite{you2026v2x} introduces large vision-language models (VLMs) into cooperative autonomous driving, transforming multi-agent coordination from reactive feature sharing to proactive semantic reasoning. By fusing vehicle-mounted camera views, infrastructure sensor data, and textual position embeddings within a unified VLM, V2X-VLM enables agents to communicate explicit driving intentions, reasoning rationales, and coordination decisions in natural language, establishing a highly interpretable medium for collective decision-making that significantly enhances safety and operational throughput in dense urban environments. These end-to-end approaches complement the modular architectures discussed above by demonstrating that the perception, communication, and planning stages can be jointly optimized within a single differentiable pipeline. Beyond single-scene reasoning, vision-language-action (VLA) models extend the cooperative-driving pipeline to the actuation layer, mapping shared perceptual tokens directly to decentralized control outputs so that perception, communication, and planning are optimized inside a single pre-trained policy~\cite{shriram2025towards}. In heterogeneous V2X fleets, natural-language intent offers a modality-agnostic lingua franca that survives sensor asymmetry between, for example, a camera-equipped micro-car and an infrastructure-side LiDAR node. The binding constraint in this driving setting is latency: autoregressive decoding of a multi-billion-parameter backbone is incompatible with the sub-second control loop required for collision avoidance, so deployable designs relegate language-level reasoning to non-time-critical layers such as mission replanning while a lightweight certified controller retains actuation authority. The broader question of how LLM-mediated multi-agent reasoning and hallucination control should be governed is taken up as a frontier topic in Section~\ref{sec:open}.

\noindent\textbf{MARL Algorithm Families for Collaborative Embodiment.} The choice of MARL algorithm significantly shapes the embodied behavior of the swarm~\cite{research0064}. Centralized training with decentralized execution (CTDE) is the dominant paradigm, as it allows agents to access global information during training while acting on local observations at deployment. Actor-critic methods such as MAPPO (multi-agent PPO)~\cite{yu2022surprising} extend single-agent proximal policy optimization to multi-agent settings by equipping each agent with a decentralized actor and a centralized critic, and have subsequently been applied to continuous-control cooperative tasks such as formation flight and coordinated navigation, complementing formal-methods-based multi-agent coordination under collaborative task specifications~\cite{research0337}. Value-based methods such as QMIX~\cite{rashid2020monotonic} factorize the joint action-value function into per-agent utilities with a monotonic mixing network, enabling efficient decentralized execution in discrete action spaces, which makes them well-suited for intersection negotiation and channel access scheduling. MADDPG (multi-agent deep deterministic policy gradient)~\cite{lowe2017multi} extends DDPG to multi-agent domains with centralized critics, and has subsequently been applied to UAV coverage optimization and active perception. In practice, the algorithm choice depends on the action space (continuous vs.\ discrete), the degree of inter-agent dependency, and the communication topology: actor-critic methods generally scale better to continuous high-dimensional action spaces, while value-decomposition methods offer stronger convergence guarantees in structured cooperative tasks.
	
\noindent\textbf{Perceptual Credit Assignment in Embodied Tasks.}
A major challenge in embodied synergy is the credit assignment that determines which agent's observation is decisive for a successful collective action. Modern GCI motifs utilize counterfactual gradients~\cite{foerster2018counterfactual} to reward agents whose shared data significantly shift the collective policy toward a safer or more efficient outcome~\cite{fu2024closely}. This mechanism incentivizes agents to focus their sensing and movement on bottleneck regions of the environment, ensuring that collective actuation remains robust when individual agents have limited local visibility.

\subsection{Shared Intent and Predictive Coordination}

An embodied swarm does not merely react to the current state; it collaboratively predicts future states to pre-emptively resolve conflicts and navigate dynamic obstacles~\cite{luo2023jfp}.

\noindent\textbf{Joint Future Prediction and Proactive Coordination.} By sharing intent tokens within the shared latent manifold~\cite{wang2025cmp}, agents can perform joint future prediction. Instead of predicting others' movements as independent stochastic processes, the swarm models the environment as a coupled dynamic system~\cite{ren2024collaborative}. This allows for proactive coordination: an agent kilometers away can adjust its trajectory based on the predicted congestion or hazards reported by the swarm's vanguard, reducing energy consumption and mission time~\cite{hu2024simulation, li2022cooperative}.

\noindent\textbf{Generative Predictive Maintenance Under Disconnection.} When the communication link is severed, the embodied CI utilizes generative world models to maintain synergy. The agents use learned predictive models to anticipate the missing neighbors' actions based on historical trajectories and shared mission priors~\cite{zhou2025v2xpnp}. This ensures a level of graceful degradation where the swarm continues its coordinated maneuver using predicted virtual peers until physical contact is re-established. This predictive loop is vital for operations in GPS-denied or high-interference zones where constant synchronization is impossible~\cite{yue2022aerial}. When such world models are grounded in large-scale multi-modal pre-training, they additionally supply language-grounded commonsense priors that fill perceptual gaps no single sensor can resolve: a disconnected ground agent can query its internal model for the most likely class and affordance of an occluded obstacle inferred from partial cues, or roll out counterfactual futures to evaluate alternative maneuvers without live peer confirmation~\cite{fu2025generative}. This reframes the world model not merely as a trajectory predictor but as a generative simulator that lets each agent rehearse coordinated actions against hypothesized swarm states, narrowing the gap between the disconnected sub-teams' internal beliefs until physical synchronization is restored.

\noindent\textbf{Topographic-Tactical Coupling in Heterogeneous Teams.} The action-perception loop is particularly powerful in heterogeneous teams where different agents occupy distinct physical niches. In these systems, synergy is achieved through a functional division of labor. Aerial agents (UAVs) provide a topographic perspective that serves as the long-range intent for the swarm, while ground agents (unmanned ground vehicles, UGVs) provide high-resolution tactical data for short-range execution. The action-perception loop facilitates a hierarchy of control: the UAV's wide-area perception dictates the UGV's path-finding, while the UGV's close-range encounters with obstacles update the UAV's global occupancy map. This multi-layered synergy allows the team to navigate environments that are far too complex for any homogeneous swarm to handle, such as dense forests or multi-level urban ruins~\cite{zhou2024enhanced, yue2022aerial}.

\subsection{Synthesis}

Embodied synergy closes the loop between collective perception and multi-agent actuation, enabling swarms to actively shape their environment, resolve sensing uncertainties through strategic movement, and execute complex missions with high resilience. Key open challenges include scaling MARL policies to large agent populations without divergence, ensuring safety guarantees during emergent coordinated maneuvers, and bridging the gap between simulation-trained policies and real-world deployment with hardware-in-the-loop validation. This bidirectional coupling of ``seeing'' and ``doing'' represents a critical dimension of the GCI roadmap outlined in Section~\ref{sec:open}. The realizability gap is sharpest where embodied synergy meets foundation models: end-to-end VLA policies that jointly output perception and control are hard to certify, their emergent coordinated maneuvers are not yet covered by formal safety analysis, and the compute budget for onboard inference on a micro-UAV is typically 10--30~W, far below what a full multi-billion-parameter backbone demands. Practical near-term deployments therefore favor a layered architecture in which a lightweight, certified reactive controller handles safety-critical actuation while a shared foundation model supplies only higher-level intent and scene understanding, keeping the unverifiable components outside the safety loop.

\begin{figure}[t]
	\begin{center}
		\includegraphics[width=\textwidth]{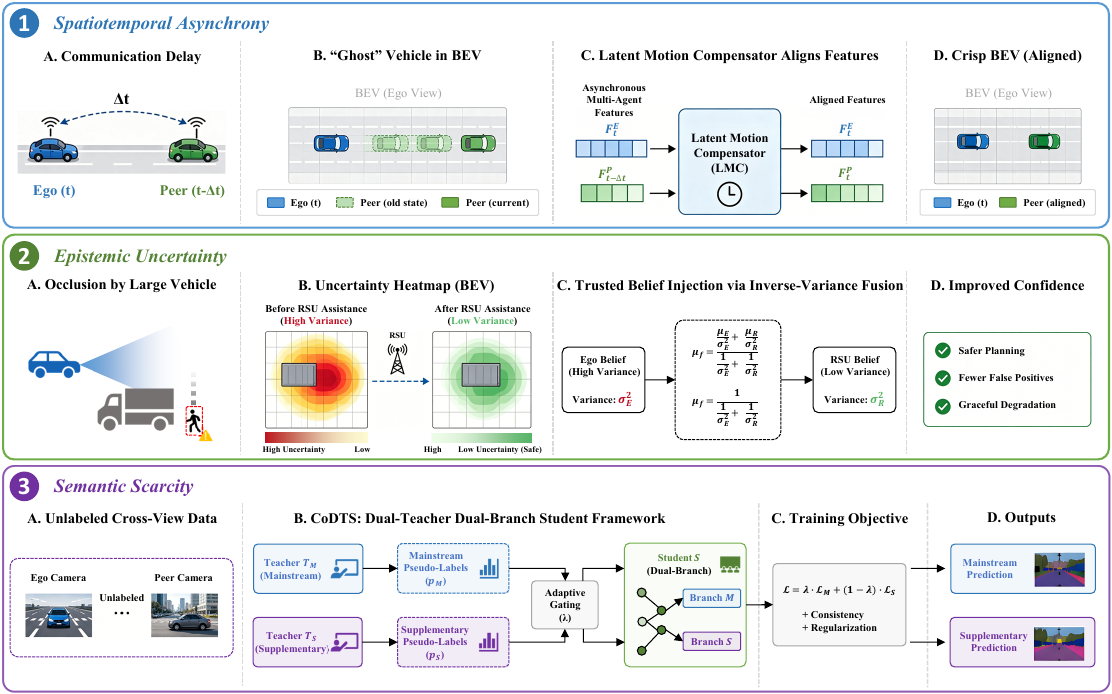}
	\end{center}
	\caption{\textbf{The ``three shadows'' of distributed collaborative perception and their corresponding structural resilience mechanisms.} 
		The taxonomy addresses three foundational systemic vulnerabilities in collective intelligence: 
		\textbf{(1) Spatiotemporal Asynchrony (The Temporal Shadow):} Communication latency $\Delta t$ and coordinate misalignments inherently induce catastrophic ghosting artifacts and localization degradation. This is mitigated through proactive neural manifold warping or predictive motion compensation networks that dynamically project stale historical features onto the current temporal baseline. 
		\textbf{(2) Epistemic Uncertainty (The Spatial Shadow):} Localized line-of-sight occlusions (e.g., dynamic obstacles or blind spots) result in high-variance, uncertified feature volumes. The framework counteracts this by quantifying evidential uncertainty distributions and performing inverse-variance-weighted belief consensus to neutralize perceptual blind spots. 
		\textbf{(3) Semantic Scarcity (The Annotation Shadow):} The high cost of multi-view cooperative annotation creates a severe training bottleneck. This is structurally resolved via the Dual Teacher-Student (CoDTS) paradigm, which distills mainstream and supplementary pseudo-labels from massive unlabeled data streams to achieve label-efficient collaborative optimization.}
	\label{fig:three_shadows_resilience}
\end{figure}

\section{Frontiers of Resilience}\label{sec:resilience}

This section surveys spatiotemporal synchronization, uncertainty quantification, and label-efficient learning as the three pillars of resilient CI deployment. The representative methods and their synergy coverage are cataloged in Table~\ref{tab:taxonomy_resilience}. These mechanisms span Dimension~4 (Learning Strategy) and safeguard \Cone\ (Semantic Disambiguation), ensuring collaboration resolves rather than amplifies entity identity under real-world degradation.

\begin{table}[t]
	\centering
	\begin{threeparttable}
		\caption{\textbf{Taxonomy of resilience methods and cognitive synergy conditions.}}
		\label{tab:taxonomy_resilience}
		\renewcommand{\arraystretch}{1.5}
		\tiny
		\setlength{\tabcolsep}{4pt}
		\renewcommand{\tabularxcolumn}[1]{m{#1}}
		\sffamily
		\begin{tabularx}{\textwidth}{
				>{\raggedright\arraybackslash\hsize=1.3\hsize}X
				>{\centering\arraybackslash\hsize=0.7\hsize}X
				>{\centering\arraybackslash\hsize=0.8\hsize}X
				>{\raggedright\arraybackslash\hsize=1.4\hsize}X
				>{\centering\arraybackslash\hsize=0.9\hsize}X
				>{\raggedright\arraybackslash\hsize=1.5\hsize}X
				>{\centering\arraybackslash\hsize=0.7\hsize}X
				>{\centering\arraybackslash\hsize=0.7\hsize}X}
			\toprule
			\textbf{Method} & \textbf{Year} & \textbf{Stage} & \textbf{Comm.\ Paradigm} & \textbf{Architecture} & \textbf{Learning Strategy} & \textbf{Application} & \textbf{Synergy} \\
			\midrule
			\rowcolor{gray!6} ROBOSAC~\cite{li2023among} & \ygray{2023} & \InterBadge & Byzantine-aware & Decentr. & Supervised + Robust & \VTwoX & \Cone \\
			GevBEV~\cite{yuan2023generating} & \ygray{2023} & \InterBadge & Periodic + EDL & Centr. & EDL + Supervised & \VTwoX & \ConeTwo \\
			\rowcolor{gray!6} MRCNet~\cite{hong2024multi} & \ygray{2024} & \InterBadge & On-demand (motion) & Decentr. & Supervised + GRU & \VTwoX & \Cone \\
			V2X-INCOP~\cite{ren2024interruption} & \ygray{2024} & \InterBadge & Predictive recovery & Decentr. & Supervised + KD & \VTwoX & \Cone \\
			\rowcolor{gray!6} MOT-CUP~\cite{su2024collaborative} & \ygray{2024} & \InterBadge & Saliency (uncertainty) & Decentr. & EDL + Bayesian & \VTwoX & \ConeTwo \\
			MA-SUE~\cite{su2024makes} & \ygray{2024} & \InterBadge & Periodic & Decentr. & SSL (Contrastive) & \VTwoX & \Cone \\
			\rowcolor{gray!6} Collab.~NeRF~\cite{zhao2024distributed} & \ygray{2024} & \InterBadge & Periodic & Decentr. & SSL (Photometric) & \UxS & \Cone \\
			CoDTS~\cite{han2025codts} & \ygray{2025} & \InterBadge & Periodic & Decentr. & Semi-supervised (KD) & \VTwoX & \Cone \\
			\bottomrule
		\end{tabularx}
		\smallskip
		\parbox{\linewidth}{\footnotesize \textit{Abbreviations:} LMC = Latent Motion Compensation, GRU = Gated Recurrent Unit, KD = Knowledge Distillation, EDL = Evidential Deep Learning, SSL = Self-Supervised Learning, Decentr.\ = Decentralized, Centr.\ = Centralized. \textit{Synergy Conditions:} \Cone\ = Semantic Disambiguation, \Ctwo\ = Pragmatic Exchange, \Cthree\ = Proactive Foraging.}
	\end{threeparttable}
\end{table}

High-entropy real-world deployments necessitate a fundamental shift from static data fusion to \textbf{resilient cyber-physical synthesis}. In the open world, CI confronts extreme perceptual entropy induced by the ``three shadows'' of distributed systems (see Figure~\ref{fig:three_shadows_resilience}): \textbf{spatiotemporal asynchrony}~\cite{lei2022latency}, \textbf{epistemic uncertainty}~\cite{su2024collaborative}, and \textbf{semantic scarcity}~\cite{su2024makes}; the latter is a domain-general challenge that is amplified in collaborative settings by the combinatorial explosion of multi-agent, multi-view annotation, rather than a collaboration-specific shadow per se. Three interlocking mechanisms underpin resilient CI. Spatiotemporal synchronization, which encompasses neural manifold warping and latent motion compensation, aligns distributed observations across space and time. Uncertainty quantification serves as the architecture of trust (including Byzantine-robust aggregation). Label-efficient learning, including consensual contrastive learning and domain generalization, opens the path toward autonomous semantic evolution when annotations are scarce. Each mechanism is examined in turn below.

\subsection{Spatiotemporal Synchronization}

CI is inherently sensitive to the stochastic jitters of the physical and network layers. In high-mobility scenarios, such as V2X or high-speed UAV swarms, the latency can result in a multi-meter spatial deviation, rendering naive fusion catastrophic~\cite{lei2022latency}.

\noindent\textbf{Neural Manifold Warping and Pose-Noise-Awareness.} Geometric offsets between agents induced by individual odometric drift, IMU bias, or GPS-denied conditions often result in ghosting artifacts where the fused manifold contains duplicate or misaligned entities. Modern architectures, such as V2X-ViT~\cite{xu2022v2x} and CoBEVT~\cite{xu2023cobevt}, depart from rigid geometric registration, employing pose-noise-aware training (PNAT), which treats spatial calibration as a learned, soft-constrained optimization problem. By injecting synthetic spatial noise during the learning phase, the neural manifold develops an inherent spatial smoothness, allowing transformer-based attention mechanisms to conceptually align features when coordinate frames are misaligned by several meters. To resolve the non-linearities of odometric drift, recent works utilize spatial transformer network (STN) based intermediate modules~\cite{jaderberg2015spatial} that predict a homography or affine transformation matrix in the feature space, aligning the BEV representations of neighbors with the ego-vehicle's frame through end-to-end backpropagation. Complementing these training-time and alignment-based strategies, CoAlign~\cite{lu2023coalign} has become a canonical framework for pose-error-robust intermediate fusion: it models an agent-object pose graph to enhance cross-agent pose consistency and applies multi-scale data fusion, correcting unknown localization errors without requiring ground-truth poses during training.

\noindent\textbf{Temporal Latent Motion Compensation (LMC).} Temporal discrepancies caused by network congestion or heterogeneous processing latencies are a primary source of fusion error. Rather than fusing stale information, modern frameworks utilize LMC to predict the future of received features. By utilizing a velocity-aware warping module~\cite{wei2023asynchrony}, agents project received feature tensors into the current ego-timestamp, ensuring that the collective consensus is synchronized in the time domain. Frameworks such as SyncNet~\cite{lei2022latency} utilize optical flow or latent flow estimation to reconstruct missing temporal intermediate states, ensuring that the fused world model remains a high-fidelity representation of the $t$-now state rather than a blurred mosaic of the $t$-past.

\noindent\textbf{Robust Communication Under Interruptions and Packet Loss.} MRCNet~\cite{hong2024multi} employs multi-scale robust fusion with GRU-based motion compensation for feature recovery during dropout; V2X-INCOP~\cite{ren2024interruption} uses communication-adaptive spatial-temporal prediction with knowledge distillation, achieving cooperative gains exceeding 12\% across three benchmarks under high packet loss.

\subsection{Uncertainty Quantification}

A critical capability for resilient CI is the ability to quantify what is not known. To prevent the poisoning of the global world model by a faulty, spoofed, or severely occluded agent, CI frameworks move beyond naive weighted averaging toward trust-centric Bayesian aggregation.

\noindent\textbf{Bayesian Epistemic Propagation and Conformal Prediction.} Instead of sharing point estimates that lack a measure of reliability, agents share probabilistic distributions. For a predictive model with parameters \(\boldsymbol{\theta}\) trained on dataset \(\mathcal{D}\), the total predictive uncertainty at a single query input \(\mathbf{x}^*\) decomposes into aleatoric and epistemic components as in Eq.~\ref{eq:uncertainty_decomp} (we retain the standard per-input Bayesian notation here: \(\mathbf{x}^*\) is a single query, \(y^*\) its model prediction, consistent with the world-state estimate \(\hat{\mathbf{Y}}\) of Section~\ref{sec:foundations} aggregated over all queries):
\begin{equation}
	\underbrace{\mathrm{Var}[y^* \mid \mathbf{x}^*, \mathcal{D}]}_{\text{Total Uncertainty}} \;=\; \underbrace{\mathbb{E}_{\boldsymbol{\theta}}[\mathrm{Var}[y^* \mid \mathbf{x}^*, \boldsymbol{\theta}]]}_{\text{Aleatoric (sensor noise, irreducible)}} \;+\; \underbrace{\mathrm{Var}_{\boldsymbol{\theta}}[\mathbb{E}[y^* \mid \mathbf{x}^*, \boldsymbol{\theta}]]}_{\text{Epistemic (model ignorance, reducible)}}.
	\label{eq:uncertainty_decomp}
\end{equation}
In collaborative settings, epistemic uncertainty concentrates in regions occluded from the ego-agent but visible to neighbors, providing a principled basis for inverse-variance-weighted Bayesian aggregation~\cite{su2024collaborative,yuan2023generating}. Conformal prediction layers add statistical coverage guarantees~\cite{mei2026perceive}.

\noindent\textbf{Adversarial Robustness and Error Isolation.} In contested environments, a single compromised agent can inject phantom obstacles to paralyze the swarm. By monitoring the consensus entropy, the swarm can detect agents whose reported features significantly deviate from the collective belief manifold. CI systems utilize M-estimators and robust loss functions (e.g., Huber loss or Tukey loss) to automatically down-weight or quarantine outlier agents. Drawing from distributed computing theory, Byzantine-robust aggregation methods originally developed for distributed gradient SGD (e.g., Krum or Median-based fusion)~\cite{blanchard2017machine} are being adapted to the neural feature space, ensuring that as long as a sufficient majority of agents are functional, the collective world model remains anchored in reliable perceptual evidence~\cite{hurl2020trupercept, gamerdinger2024robust, li2023among}.

\subsection{Label-Efficient Learning}

The annotation bottleneck lies in the immense cost of manually labeling multi-agent, multi-modal datasets. This remains a primary barrier to scaling collaborative intelligence.

\noindent\textbf{Consensual Contrastive Learning (CCL).} By exploiting the inherent redundancy of overlapping fields of view, agents can treat their neighbors' perspectives as pseudo-labels for their own occluded views. MA-SUE~\cite{su2024makes} utilizes contrastive loss to align features from different agents observing the same scene. For instance, if Agent A sees the front of a car and Agent B sees the back, the CCL objective forces the latent representations of both agents to be consistent with a unified 3D object concept, effectively teaching each agent to infer the complete structure of an object from a partial view without any human intervention. Recent advances in collaborative neural radiance field (NeRF) approaches~\cite{zhao2024distributed} or Gaussian Splatting~\cite{zeng2025multi} allow agents to build a high-fidelity 3D radiance field collectively, where the photometric consistency between distributed viewpoints acts as a strong supervisory signal, enabling the swarm to learn geometry and semantics through the sheer volume of distributed observations.

\noindent\textbf{Dual Teacher-Student Frameworks.} CoDTS~\cite{han2025codts} introduces a dual teacher-student paradigm with static and dynamic pseudo-label generation, maintaining robust detection with as few as 10\% labeled data across V2X-Sim and OPV2V, demonstrating label-efficient collaborative perception without reliance on fully supervised training.

\noindent\textbf{Domain Generalization and the ``Sim-to-Real'' Bridge.} To overcome the scarcity of real-world edge cases (e.g., accidents or extreme weather), researchers utilize high-fidelity simulated environments such as V2X-Sim~\cite{li2022v2x}. Heterogeneous teams comprising high-end infrastructure, mid-tier autonomous vehicles, and low-cost micro-drones suffer from feature-level divergence; recent works~\cite{hu2024toward} introduce domain adversarial neural networks (DANN)~\cite{chen2018domain} to reconcile these discrepancies, ensuring semantic equivalence where the representation of a pedestrian remains invariant regardless of whether the data comes from a 128-beam LiDAR or a low-resolution monocular camera. Furthermore, by using diffusion models to synthesize adversarial weather (e.g., heavy fog or snow) on top of clear-day data, the swarm can be pre-trained to maintain perceptual stability in hostile environments~\cite{li2025v2x}.

\subsection{Synthesis}

Spatiotemporal synchronization, uncertainty quantification, and label-efficient learning, the three frontiers surveyed above, collectively advance CI from brittle, sensor-sharing configurations toward resilient, self-healing architectures. By integrating uncertainty-aware fusion with neural manifold warping, future GCI systems can maintain identity persistence and semantic clarity even in contested, degraded, and unpredictable operational environments. Label-efficient self-supervision closes the loop by reducing dependency on costly multi-agent annotations. Together, these mechanisms form the resilience backbone of the GCI roadmap, as organized in Table~\ref{tab:three_shadows}.

\begin{table}[t]
	\centering
	\begin{threeparttable}
		\caption{\textbf{The ``Three Shadows'' of collaborative resilience: systemic vulnerabilities, mechanisms, and open challenges.} Each shadow exposes a distinct failure surface in distributed perception; the corresponding mechanisms represent the current state of the art in addressing them.}
		\label{tab:three_shadows}
		\renewcommand{\arraystretch}{1.4}
		\tiny
		\setlength{\tabcolsep}{3pt}
		\renewcommand{\tabularxcolumn}[1]{m{#1}}
		\sffamily
		
		\begin{tabularx}{\textwidth}{
				>{\hsize=0.6\hsize\raggedright\arraybackslash}X 
				>{\hsize=0.8\hsize\raggedright\arraybackslash}X 
				>{\hsize=1.1\hsize\raggedright\arraybackslash}X 
				>{\hsize=1.2\hsize\raggedright\arraybackslash}X 
				>{\hsize=1.3\hsize\raggedright\arraybackslash}X
			}
			\toprule
			\centering\textbf{Systemic Vulnerability} & 
			\centering\textbf{Root Cause} & 
			\centering\textbf{Key Mechanisms} & 
			\centering\textbf{Representative Methods} & 
			\centering\arraybackslash\textbf{Open Challenges} \\
			\midrule
			
			Spatiotemporal Asynchrony & 
			Communication latency, heterogeneous clocks, coordinate misalignment & 
			Neural manifold warping (PNAT), latent motion compensation (LMC), motion-aware robust communication & 
			V2X-ViT~\cite{xu2022v2x}, CoBEVT~\cite{xu2023cobevt}, SyncNet~\cite{lei2022latency}, MRCNet~\cite{hong2024multi}, V2X-INCOP~\cite{ren2024interruption} & 
			Real-time inference under ms latency budgets; formal alignment guarantees for safety-critical settings \\
			
			\cellcolor{gray!6} Epistemic Uncertainty & 
			\cellcolor{gray!6} Sensor degradation, adversarial agents, occlusion & 
			\cellcolor{gray!6} Evidential deep learning (EDL), conformal prediction, inverse-variance Bayesian aggregation, Byzantine-robust fusion & 
			\cellcolor{gray!6} MOT-CUP~\cite{su2024collaborative}, GevBEV~\cite{yuan2023generating}, ROBOSAC~\cite{li2023among}, Krum/Median~\cite{blanchard2017machine} & 
			\cellcolor{gray!6} Provable Byzantine resilience with formal convergence guarantees; runtime uncertainty calibration on embedded hardware \\
			
			Semantic Scarcity & 
			High multi-view annotation cost, domain shift across deployments & 
			Consensual contrastive learning (CCL), dual teacher-student (CoDTS), domain adversarial adaptation (DANN) & 
			MA-SUE~\cite{su2024makes}, CoDTS~\cite{han2025codts}, domain generalization~\cite{hu2024toward} & 
			Closing the sim-to-real generalization gap; scaling to diverse real-world edge cases with minimal supervision \\
			\bottomrule
		\end{tabularx}
	\end{threeparttable}
\end{table}

\section{Domain-Specific Synergy}\label{sec:domain}

The synergy methods surveyed in this section and their mapping to the five-dimensional taxonomy and C1--C3 conditions are listed in Table~\ref{tab:taxonomy_domain}.

\begin{figure}[t]
	\begin{center}
		\includegraphics[width=\linewidth]{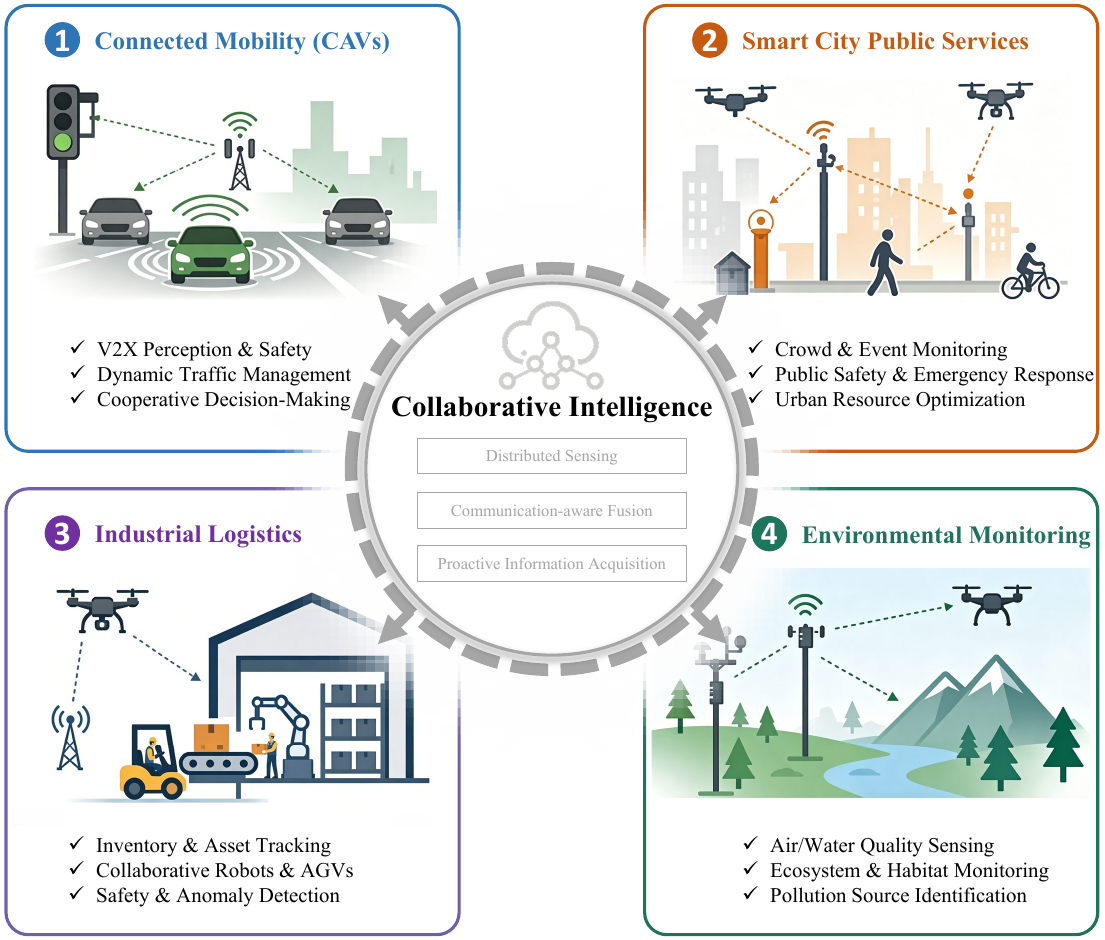}
	\end{center}
	\caption{\textbf{Domain-specific synergy across collaborative-intelligence deployments.} The illustration links four application contexts (industrial logistics, connected mobility, smart-city public services, and environmental monitoring) to the shared requirement for distributed sensing, communication-aware fusion, and proactive information acquisition.}
	\label{fig:domain_synergy}
\end{figure}

\begin{table}[t]
	\centering
	\begin{threeparttable}
		\caption{\textbf{Taxonomy of domain-specific methods and cognitive synergy conditions.}}
		\label{tab:taxonomy_domain}
		\renewcommand{\arraystretch}{1.5}
		\tiny
		\setlength{\tabcolsep}{4pt}
		\renewcommand{\tabularxcolumn}[1]{m{#1}}
		\sffamily
		\begin{tabularx}{\textwidth}{
				>{\raggedright\arraybackslash\hsize=1.3\hsize}X
				>{\centering\arraybackslash\hsize=0.7\hsize}X
				>{\centering\arraybackslash\hsize=0.8\hsize}X
				>{\raggedright\arraybackslash\hsize=1.4\hsize}X
				>{\centering\arraybackslash\hsize=0.9\hsize}X
				>{\raggedright\arraybackslash\hsize=1.5\hsize}X
				>{\centering\arraybackslash\hsize=0.7\hsize}X
				>{\centering\arraybackslash\hsize=0.7\hsize}X}
			\toprule
			\textbf{Method} & \textbf{Year} & \textbf{Stage} & \textbf{Comm.\ Paradigm} & \textbf{Architecture} & \textbf{Learning Strategy} & \textbf{Application} & \textbf{Synergy} \\
			\midrule
			Collab.~SLAM~\cite{zou2019collaborative} & \ygray{2019} & \InterBadge & Periodic & Decentr. & SLAM & \UxSLog & \Cone \\
			\rowcolor{gray!6} MRSLAM~\cite{hernandez2020real} & \ygray{2020} & \InterBadge & Periodic & Decentr. & SLAM + Supervised & \UxSLog & \Cone \\
			BM2CP~\cite{zhao2023bm2cp} & \ygray{2023} & \InterBadge & Periodic & Edge (RSU) & Supervised & \VTwoX & \ConeTwo \\
			\rowcolor{gray!6} HM-CoPept~\cite{zha2025heterogeneous} & \ygray{2025} & \InterBadge & Periodic & Hybrid & Supervised & \VTwoX & \ConeTwo \\
			AccBEV~\cite{shi2025v2v} & \ygray{2025} & \InterBadge & Gen.\ Reconstr.\ (CVAE) & Decentr. & Supervised + Gen. & \VTwoX & \ConeTwo \\
			\rowcolor{gray!6} V2X-DGW~\cite{li2025v2x} & \ygray{2025} & \InterBadge & Periodic & Hybrid & Supervised + DG & \VTwoX & \Cone \\
			\bottomrule
		\end{tabularx}
		\smallskip
		\parbox{\linewidth}{\footnotesize \textit{Abbreviations:} RSU = Roadside Unit, CVAE = Conditional Variational Auto-Encoder, Gen.\ = Generative, DG = Domain Generalization, Decentr.\ = Decentralized. \textit{Synergy Conditions:} \Cone\ = Semantic Disambiguation, \Ctwo\ = Pragmatic Exchange, \Cthree\ = Proactive Foraging.}
	\end{threeparttable}
\end{table}

The transition from deterministic data fusion to adaptive cognitive synergy is crystallized across four primary strategic domains: connected and automated vehicles (CAVs), UxS, industrial logistics, and smart cities, as illustrated in Figure~\ref{fig:domain_synergy}. V2X is the only domain with routine cross-method comparison; UxS, industrial logistics, and smart-city CI remain at earlier maturity stages.
\subsection{Connected and Automated Vehicular Networks (V2X)}

Autonomous driving represents the most mature and safety-critical application of collaborative intelligence. The V2X perception field benefits from a rich ecosystem of standardized, openly available benchmarks (OPV2V~\cite{xu2022opv2v}, DAIR-V2X~\cite{yu2022dair}, V2X-Sim~\cite{li2022v2x}, V2V4Real~\cite{xu2023v2v4real}, and V2X-Seq~\cite{yu2023v2x}), with over a dozen methods evaluated under comparable conditions, making it the only CI domain where rigorous cross-method comparison on shared benchmarks is routinely feasible. The integration of vehicle-to-vehicle (V2V) and vehicle-to-infrastructure (V2I) communication creates a networked perception system that extends beyond the physical and computational limits of individual onboard sensor suites~\cite{bai2024survey}, a direction extended by parallel-driving foundation-model frameworks~\cite{research0349} and multidimensional self-driving interactive-cognition models~\cite{research0903}.

\noindent\textbf{Occlusion Resiliency and Perceptual Horizon Extension.} In high-density urban environments, failures in ego-centric perception are frequently induced by the dynamic occlusion paradox, where moving large-scale actors (e.g., buses, trucks) hide safety-critical entities. CI mitigates these risks by reconstructing the neural manifold from non-overlapping, distributed viewpoints. To maintain robust situational awareness during sensor degradation, BM2CP~\cite{zhao2023bm2cp} and HM-CoPept~\cite{zha2025heterogeneous} utilize high-fidelity geometric priors from fixed roadside infrastructure to guide the feature extraction of mobile agents, enabling vehicles with low-cost sensor suites to achieve quasi-LiDAR spatial reasoning by borrowing the geometric certainty of roadside units (RSUs). To optimize the communication-fidelity trade-off, architectures such as Fusion2comm~\cite{chu2025occlusion} dynamically prioritize bandwidth for occluded foreground features, while systems like AccBEV~\cite{shi2025v2v} employ conditional variational auto-encoders (CVAE) and normalizing flows to synthesize shared features in the BEV manifold, effectively extending the perception horizon beyond the vehicle's physical line-of-sight~\cite{ngo2023cooperative, liu2023region}. Occluded object detection (see also the robustness discussion in Section~\ref{sec:resilience}) remains an active area, with dedicated reviews cataloging methods that leverage temporal consistency, contextual reasoning, and multi-view fusion to recover objects hidden behind dynamic obstacles~\cite{ruan2023review}. Real-world V2X communication also suffers from asynchronous clocks and network jitter; the spatiotemporal synchronization techniques detailed in Section~\ref{sec:resilience} (e.g., latent motion compensation and velocity-aware feature warping) are especially critical in high-mobility V2X scenarios, where millisecond-level delays translate to meter-level spatial errors at highway speeds.

\noindent\textbf{Intersection Safety and Infrastructure-Augmented Coordination.} Unsignalized intersections represent high-risk zones where occlusions and unpredictable pedestrian movements converge into dangerous corner cases~\cite{su2025occlusion, bejarbaneh2024exploring}. RSUs act as distributed anchor nodes, performing data association across heterogeneous sensor streams to resolve trajectory inconsistencies in real-time~\cite{zhang2023roadside, liu2025rsu}, and provide an augmented global view that utilizes historical traffic flow priors and multi-modal fusion to offer localized driving strategies that preemptively resolve potential conflicts~\cite{wang2024augmented, xiao2023perception}. To ensure system continuity during RSU downtime or network partitions, decentralized strategies leverage peer-to-peer V2V coordination. By sharing intent-encoded latent features, vehicles collaboratively optimize passing orders and speed profiles, shifting from reactive braking to proactive flow optimization and creating a fluid traffic state where throughput is maximized without the need for traditional signaling~\cite{pan2025cooperative, pei2021distributed}.

\noindent\textbf{Resilient Platooning and Adverse Weather Perception.} Platooning uses high-speed V2V synchronization to bypass human reaction latencies, drastically reducing aerodynamic drag and energy consumption~\cite{hu2024simulation, li2022cooperative}. This synergy is most critical during adverse weather operations (e.g., heavy rain, dense fog, or snow), where individual sensors suffer from severe signal scattering and optical attenuation~\cite{nanda2025influence, kou2025adverse}. CI provides a redundant and weather-invariant informational layer: frameworks such as V2X-DGW~\cite{li2025v2x} utilize adversarial domain adaptation to extract weather-proof features, correlating LiDAR returns with camera-based semantics from multiple agents to maintain clear collective sight when individual sensors are effectively blinded. Furthermore, systems can integrate uncertainty quantification, where each agent attaches a confidence score to its shared data based on localized weather intensity, allowing the fusion node to dynamically weight the most reliable sensor sources~\cite{ning2025coinfra}.

\subsection{Unmanned Aerial and Heterogeneous Systems (UxS)}

Unmanned aerial systems (UAS) introduce a critical vertical dimension to the collaborative manifold, acting as rapidly deployable aerial sensors that provide a topographic perspective difficult for ground agents to obtain~\cite{fan2023area, ahmad2025future}. Compared with the V2X domain, the UxS collaborative perception literature is smaller in volume and less standardized: several collaborative-native benchmarks have recently emerged (including CoPerception-UAV~\cite{hu2022where2comm}, U2UData~\cite{feng2024u2udata}, UAV3D~\cite{ye2024uav3d}, Griffin~\cite{wang2026griffin}, AGC-Drive~\cite{hou2026agc}, CoPeD~\cite{zhou2024coped}, and AirCopBench~\cite{zha2026aircopbench}), yet reported results are still derived mostly from single-study simulation validations rather than multi-team benchmark comparisons. The discussion below therefore reflects the state of an emerging rather than a mature subfield.

\noindent\textbf{Proactive Surveillance and Search-and-Rescue.} UAV swarms utilize superior maneuverability to access GPS-denied or hazardous disaster zones where ground access is impossible~\cite{fan2023area, ahmad2025future}. In search-and-rescue (SAR) missions, timely information acquisition is the decisive factor for survival: integrated air-ground networks, in which UGVs collect high-resolution ground data while UAVs provide wide-area detection, represent an effective form of heterogeneous synergy~\cite{hao2024rule, research0920}, and swarms utilize collaborative SLAM to map debris and locate victims simultaneously under a unified coordinate frame~\cite{ngo2022uav}. In the active perception paradigm, agents do not merely observe; they co-optimize perception and actuation using multi-agent deep reinforcement learning (MADRL), with aerial agents dynamically adjusting their flight paths to minimize global estimation entropy and strategically positioning themselves to peek behind rubble or through windows to maximize collective information gain~\cite{zhou2024enhanced, nguyen2025csaot}.

\noindent\textbf{Precision Industrial Inspection and High-Resolution Photogrammetry.} Collaborative UAS perform high-resolution photogrammetry and multi-spectral remote sensing for industrial automation~\cite{li2022intelligent}. For critical infrastructure such as high-voltage powerlines and bridges, agents collaboratively perform 3D reconstruction, eliminating the self-occlusion inherent in complex geometric structures by sharing viewpoints in real-time~\cite{zhang2024soar}. In precision agriculture, swarms exchange multi-spectral maps to monitor crop health, enabling synchronization of fertilization at the centimeter scale. In this scenario, one agent detects nitrogen deficiency and another autonomously triggers localized spraying, closing the perception-action loop across the entire swarm~\cite{tsouros2019review}.

\subsection{Industrial Logistics and Warehouse Robotics}

The evidence base for collaborative perception in industrial logistics remains fragmented: research is distributed across application-specific systems and proprietary industrial deployments, and standardized benchmarks comparable to OPV2V~\cite{xu2022opv2v} or DAIR-V2X~\cite{yu2022dair} do not yet exist for warehouse and logistics settings. Reported results often reflect single-vendor deployments rather than multi-team comparative evaluations. The discussion below therefore surveys emerging capabilities rather than mature, benchmark-validated systems~\cite{chen2025survey}. With this caveat, the domain shows considerable promise: in the confined and dynamic environments of smart warehouses, collaborative intelligence enables coordinated operation from isolated robots toward synchronized, self-optimizing logistics~\cite{priyanta2024towards, geetha2023autonomous}. Swarms of micro-drones and ground-based autonomous mobile robots (AMRs) share motion states and feature descriptors to mitigate the odometric drift that plagues single-agent SLAM in feature-poor environments (e.g., long aisles of identical shelves)~\cite{zhao2023review, zou2019collaborative}, and advanced frameworks such as MRSLAM~\cite{hernandez2020real} reconcile extreme viewpoint disparities (aerial-to-ground) to ensure globally consistent mapping in utility tunnels and automated ports~\cite{yue2022aerial, wang2025research}. Furthermore, AMRs maintain real-time communication with overhead assets to optimize sorting and inventory tracking~\cite{pratissoli2024hierarchical}. This crowd-sourced perception enables global routing optimization in large-scale smart ports and deep mines, significantly reducing idle times, fuel consumption, and operational carbon emissions. The system acts as a distributed scheduler: perception of a blockage in one sector is instantly propagated to reroute the entire fleet~\cite{xie2024conflict,geetha2023autonomous, nair2024collaborative, singh2024multi}.

\subsection{Smart Cities}

The ultimate scaling of collaborative intelligence moves beyond physical navigation toward social and environmental intelligence, where the swarm acts as the distributed nervous system of the urban environment~\cite{ahmad2025future, hamrouni2023multi}. It should be noted, however, that smart-city CI applications remain predominantly at the prototype and simulation stage, with limited large-scale, multi-city validation; the discussion below surveys emerging directions and conceptual prototypes rather than mature, benchmark-validated deployment practices.

\noindent\textbf{Crowd Management and Public Safety Ecosystems.} (\ConeTwo) UAV-based monitoring networks provide a wide-area perspective for real-time crowd density estimation and emergency evacuation planning~\cite{hamrouni2023multi}. By sharing and fusing thermal and optical data, swarms can reveal congestion levels in blind spots (e.g., behind stadium exits or subway entrances), enabling authorities to intervene with sub-minute latency during critical safety incidents. Modern urban swarms integrate Federated Learning and Differential Privacy, allowing agents to collaborate on crowd analysis without transmitting identifiable facial data, thus aligning technological utility with global privacy mandates~\cite{ahmad2025future, chhikara2021federated}.\looseness=-1

\noindent\textbf{Tiered Multi-modal Environmental Monitoring.} (\ConeTwo) CI serves as a vital tool for large-scale environmental protection and pollution tracking~\cite{chhikara2021federated}. UAV swarms perform multi-point cooperative sensing for 3D air quality mapping. Unlike static sensors, the swarm can trace the 3D plume of a gas leak in real-time, identifying localized sources that would be missed by traditional monitoring stations~\cite{chhikara2021federated2}. In maritime domains, the integration of autonomous surface vehicles (ASVs) and UAVs allows for tiered monitoring: UAVs identify pollution belts (e.g., oil spills) from the air using hyper-spectral imaging, guiding ASVs to conduct close-range, high-resolution water measurements at the exact coordinates of the plume. This hierarchical approach ensures that environmental data are both wide in scope and high-fidelity in detail, enabling rapid response to ecological threats~\cite{pinto2013collaborative,bi2024cooperative}.\looseness=-1

\noindent\textbf{Toward a Global Semantic Manifold.}
The domain-specific synergies described above will converge into a global semantic manifold. The distinction between car perception, drone mapping, and infrastructure sensing will dissolve. All autonomous agents can contribute to and draw from a shared, real-time digital twin of the physical world. This will require the resolution of significant open challenges in elastic fusion to dynamically scale its collaborative depth based on real-time mission criticality and available bandwidth. This ensures that the collective sight of the swarm remains resilient in the face of the ever-increasing complexity of human-centric environments.

\section{Trustworthy Synergy}\label{sec:trust}

Trustworthy CI requires balancing functional safety under operational uncertainty with privacy preservation during sensitive data exchange. This section examines four axes: system safety and resilience, privacy-preserving collaboration, the safety-privacy-utility trade-off, and policy-aware GCI including liability, fairness, and standardization. A taxonomy of the representative methods is provided in Table~\ref{tab:taxonomy_trust}. This section provides the trust guarantee for \Cone--\Cthree, examining the safety, privacy, and accountability mechanisms that ensure collaboration introduces no new failure modes or unresolved liability.

\begin{table}[t]
	\centering
	\begin{threeparttable}
		\caption{\textbf{Taxonomy of trustworthy CI methods and cognitive synergy conditions.} Methods are mapped onto the proposed five-dimensional framework.}
		\label{tab:taxonomy_trust}
		\renewcommand{\arraystretch}{1.5}
		\tiny
		\setlength{\tabcolsep}{4pt}
		\renewcommand{\tabularxcolumn}[1]{m{#1}}
		\sffamily
		\begin{tabularx}{\textwidth}{
				>{\raggedright\arraybackslash\hsize=1.3\hsize}X
				>{\centering\arraybackslash\hsize=0.7\hsize}X
				>{\centering\arraybackslash\hsize=0.8\hsize}X
				>{\raggedright\arraybackslash\hsize=1.4\hsize}X
				>{\centering\arraybackslash\hsize=0.9\hsize}X
				>{\raggedright\arraybackslash\hsize=1.5\hsize}X
				>{\centering\arraybackslash\hsize=0.7\hsize}X
				>{\centering\arraybackslash\hsize=0.7\hsize}X}
			\toprule
			\textbf{Method} & \textbf{Year} & \textbf{Stage} & \textbf{Comm.\ Paradigm} & \textbf{Architecture} & \textbf{Learning Strategy} & \textbf{Application} & \textbf{Synergy} \\
			\midrule
			TruPercept~\cite{hurl2020trupercept} & \ygray{2020} & \InterBadge & Trust-aware & Decentr. & Supervised + Trust & \VTwoX & \Cone \\
			\rowcolor{gray!6} Oblivious Fusion~\cite{curran2021oblivious} & \ygray{2021} & \InterBadge & Periodic (secret-shared) & Decentr.\ (SMPC) & Supervised + SMPC & \VTwoX & \Cone \\
			FedDrive~\cite{fantauzzo2022feddrive} & \ygray{2022} & \InterBadge & Periodic (gradients) & Decentr.\ (FL) & FL + Supervised & \VTwoX & \Cone \\
			\rowcolor{gray!6} Byz.-Robust~\cite{gamerdinger2024robust} & \ygray{2024} & \InterBadge & Byzantine-robust & Decentr. & Supervised + Robust & \VTwoX & \Cone \\
			FedCP~\cite{zhang2024federated} & \ygray{2024} & \InterBadge & Periodic (gradients) & Decentr.\ (FL) & FL + Supervised & \VTwoX & \Cone \\
			\rowcolor{gray!6} Secure MKHE~\cite{yang2024secure} & \ygray{2024} & \InterBadge & Periodic (encrypted) & Centr.\ (HE) & Supervised + HE & \VTwoX & \Cone \\
			\bottomrule
		\end{tabularx}
		\smallskip
		\parbox{\linewidth}{\footnotesize \textit{Abbreviations:} FL = Federated Learning, HE = Homomorphic Encryption, SMPC = Secure Multi-Party Computation, Decentr.\ = Decentralized, Centr.\ = Centralized. \textit{Synergy Conditions:} \Cone\ = Semantic Disambiguation, \Ctwo\ = Pragmatic Exchange, \Cthree\ = Proactive Foraging.}
	\end{threeparttable}
\end{table}

\subsection{System Safety and Resilience}

Ensuring the safety of a collaborative swarm is essential to its transition from controlled testbeds to open-world deployments. In the context of GCI, safety is redefined as a multilayer defense architecture that transcends individual reliability to ensure collective resilience across functional, data, and temporal dimensions.

\noindent\textbf{Functional Safety and Certified Robustness.} The integration of collaborative perception into safety-critical pipelines necessitates compliance with international safety standards, e.g., ISO 26262~\cite{ISO26262_2018} and ISO 21448 (SOTIF)~\cite{ISO21448_2019}. These standards require a rigorous quantification of safety of the intended functionality, where the swarm must mitigate risks arising from erroneous perception events, such as failing to detect a pedestrian due to shared data noise.

To provide mathematical guarantees in high-stakes environments, modern research leverages certified robustness via Randomized Smoothing~\cite{cohen2019certified}. Unlike empirical robustness, this method constructs a smoothed classifier by injecting controlled Gaussian noise into the feature-sharing process. This ensures that the collective world model remains provably consistent under bounded \(L_2\) perturbations in the input space, effectively insulating the swarm against environmental disturbances or sophisticated adversarial attacks that might otherwise induce catastrophic decision-making errors~\cite{jeong2021smoothmix}.\looseness=-1

\noindent\textbf{Data Trustworthiness and Adversarial Resilience.} In a decentralized swarm, the integrity of the shared world model is paramount~\cite{hurl2020trupercept}. The system must resolve the information credibility problem: how to distinguish between a genuine observation and a malicious or faulty signal. Trust-aware fusion frameworks address this by assigning time-varying reliability scores to each node based on latency, spatial consistency, and historical performance. To counter deliberate manipulation, such as the injection of phantom obstacles into the shared BEV manifold, the system employs methods such as ROBOSAC~\cite{li2023among}, accepting only detections confirmed by multiple independent viewpoints to filter out spoofed data and ensure that the collective belief is anchored in physical reality~\cite{gamerdinger2024robust}.

\noindent\textbf{Fail-safe Fallback, Asynchrony, and Graceful Degradation.} Runtime monitors detect temporal inconsistency from heterogeneous clocks and network jitter, triggering conservative fallback before erroneous perception propagates to the planning stack~\cite{lei2022latency}. When communication quality drops below safety-critical thresholds, DRL-optimized policies autonomously transition from collaborative to local perception mode, redistributing trust among local sensors based on epistemic uncertainty; in cases of total isolation, the agent maintains operation by relying on its local sensors~\cite{li2023learning}.

\subsection{Privacy-Preserving Collaboration}

Sharing high-dimensional data across networks risks exposing personally identifiable information (PII), such as facial features, license plates, and movement trajectories~\cite{liu2023towards, yazgan2024collaborative}.

\noindent\textbf{Data Exposure and Differential Privacy.} To satisfy global privacy mandates (e.g., GDPR), researchers are integrating differential privacy (DP) into the perception pipeline. By applying DP to shared feature maps~\cite{seif2024over} and trajectories~\cite{takagi2023geo, ma2021trajectory}, the system ensures that an adversary cannot reconstruct the specific identity of a vehicle or its passengers from the shared latent space. Furthermore, secure computation and privacy-preserving feature extraction methods~\cite{bai2022privacy} are integrated into the collaborative perception pipeline to prevent adversarial reconstruction of sensitive scene information while preserving the geometric features required for the navigation task.

\noindent\textbf{Privacy-Preserving Computation Architectures.} The shift toward GCI necessitates architectures that decouple utility from access. Federated learning (FL) for perception, exemplified by FedDrive~\cite{fantauzzo2022feddrive}, Hierarchical FL~\cite{song2022federated}, and FedDWA~\cite{zhang2024federated}, allows agents to collaborate on model training by sharing gradient updates instead of raw data. This data-local, model-global approach ensures that raw sensory streams never leave the individual agent's hardware. High-level synergy is further enabled by homomorphic encryption (HE) and secure multi-party computation (SMPC): HE allows the fusion node to aggregate features while they are still in an encrypted state, ensuring that the central unit has no access to the plain text~\cite{yang2024secure}, while SMPC splits features into secret shares so that no single participant, including the aggregator, can reconstruct the private data of another~\cite{zhao2019secure, curran2021oblivious}. Recent work extends privacy-preserving collaboration to open-world settings with unknown collaborators, where agents that have not previously registered or shared keys must still be able to contribute securely to the collective perception task~\cite{lu2025privacy}. Table~\ref{tab:privacy_techniques} provides a comparative overview of the principal privacy-preserving paradigms and their operational trade-offs.

\begin{table}[t]
	\centering
	\begin{threeparttable}
		\caption{\textbf{Comparative overview of privacy-preserving techniques for collaborative perception.} Each paradigm occupies a distinct position in the trade space between protection strength, communication overhead, and real-time feasibility.}
		\label{tab:privacy_techniques}
		\renewcommand{\arraystretch}{1.5}
		\tiny
		\setlength{\tabcolsep}{5pt}
		\renewcommand{\tabularxcolumn}[1]{m{#1}}
		\sffamily
		
		\begin{tabularx}{\textwidth}{
				>{\hsize=2.0\hsize\raggedright\arraybackslash}X 
				>{\hsize=1.0\hsize\centering\arraybackslash}X 
				>{\hsize=0.6\hsize\centering\arraybackslash}X 
				>{\hsize=0.6\hsize\centering\arraybackslash}X 
				>{\hsize=0.6\hsize\centering\arraybackslash}X 
				>{\hsize=1.2\hsize\raggedright\arraybackslash}X
			}
			\toprule
			\centering\textbf{Technique} & 
			\textbf{Protection Granularity} & 
			\textbf{Comm.\ Overhead} & 
			\textbf{Accuracy Penalty} & 
			\textbf{Real-Time} & 
			\centering\arraybackslash\textbf{Representative Work} \\
			\midrule
			
			Differential Privacy (DP) & 
			Feature / trajectory & 
			Low & 
			Moderate & 
			High & 
			Feature-map DP~\cite{seif2024over}, trajectory DP~\cite{ma2021trajectory}, DP-aware fusion~\cite{lu2025privacy} \\
			
			\cellcolor{gray!6} Federated Learning (FL) & 
			\cellcolor{gray!6} Model & 
			\cellcolor{gray!6} Moderate & 
			\cellcolor{gray!6} Low & 
			\cellcolor{gray!6} Moderate & 
			\cellcolor{gray!6} FedDrive~\cite{fantauzzo2022feddrive}, FedCP~\cite{zhang2024federated}, Hierarchical FL~\cite{song2022federated} \\
			
			Homomorphic Encryption (HE) & 
			Raw data / feature & 
			Very high & 
			None & 
			Low & 
			Secure MKHE~\cite{yang2024secure} \\
			
			\cellcolor{gray!6} Secure Multi-Party Computation (SMPC) & 
			\cellcolor{gray!6} Raw data / feature & 
			\cellcolor{gray!6} High & 
			\cellcolor{gray!6} None & 
			\cellcolor{gray!6} Low--Mod. & 
			\cellcolor{gray!6} Oblivious Fusion~\cite{curran2021oblivious}, SMPC-based fusion~\cite{zhao2019secure} \\
			\bottomrule
		\end{tabularx}
	\end{threeparttable}
\end{table}

\noindent\textbf{The Safety-Privacy-Utility Triad.}
The selection of a fusion level (see Section~\ref{sec:arch} for architectural taxonomy) is the primary driver of the balance across the reliability triangle, which we categorize into three paradigms. Raw data sharing provides the richest context for resolving occlusions but incurs prohibitive privacy risks and massive bandwidth costs, often conflicting with data minimization principles required by international regulations. Feature-level sharing represents the current state-of-the-art middle ground: by sharing compressed neural features rather than raw sensor data, agents preserve the majority of perception gains while reducing the surface area for privacy leakage, though intermediate features remain vulnerable to reconstruction attacks and therefore require explicit privacy-preserving mechanisms. Decision-level sharing offers the highest level of confidentiality by exchanging high-level metadata, but is stochastically brittle because it lacks the corrective power to resolve deep occlusions; agents cannot mutually refine their low-level perceptual hypotheses, leading to a substantial safety drop when local detectors fail.

\subsection{Policy-Aware GCI}

As GCI transitions from isolated experimental testbeds to smart cities, its deployment is no longer a functional challenge but a regulatory and ethical one. The transition to a collective architecture necessitates a new framework for policy-aware GCI, where the algorithms are not only optimized for detection accuracy or latency but are also aligned with the principles of legal accountability, algorithmic fairness, and global industrial standards.

\noindent\textbf{Liability Attribution: From Neural Fusion to Auditable Traceability.} A key legal challenge in GCI is the responsibility gap: when a fused perceptual error leads to a catastrophic decision, identifying the causal contribution of each participant in a decentralized swarm is challenging. A practical traceability layer would log the source agent ID, timestamp, pose estimate, confidence score, communication state, and the compressed feature or decision payload consumed by the fusion module, enabling post-incident forensic reconstruction. Blockchain-based frameworks~\cite{gupta2023reverse, yao2023accident} have been proposed for tamper-proof evidence preservation and reverse liability tracing in autonomous driving, though lighter authenticated logging with trust-modeling mechanisms may be more practical for latency-critical deployments~\cite{hurl2020trupercept}. The key requirement is forensic reconstructability: investigators and runtime monitors must be able to determine whether a failure was dominated by corrupted features, stale packets, pose drift, or downstream planning assumptions.

\noindent\textbf{Algorithmic Fairness and Data Sovereignty.} Collaborative intelligence must address the structural inequities that arise in heterogeneous multi-agent settings. In a swarm with uneven sensor capabilities, agents with lower-quality sensors can be marginalized during fusion, leading to systemic blind spots in specific demographic or geographic zones. GCI envisions fairness-aware fusion that prevents the global manifold from over-indexing on high-end infrastructure data at the expense of mobile agents, countering a digital divide in perception where urban districts with dense RSUs achieve higher safety levels than resource-constrained rural environments~\cite{gao2025airv2x}. Furthermore, the swarm must treat data as a sovereign asset: federated learning methods~\cite{fantauzzo2022feddrive, zhang2024federated} allow agents to contribute to the collective intelligence without forfeiting ownership of their raw sensory streams.

\noindent\textbf{Standardization Roadmap: Aligning AI Models with V2X Protocols.} The widespread adoption of GCI is contingent upon the convergence of neural architectures and telecommunication standards, as exemplified by recent transformer-based V2X perception frameworks~\cite{xu2022v2x}. The standardization gap between high-frequency AI iterations and slow-moving regulatory cycles remains a critical hurdle~\cite{han2023collaborative}. Current V2X standards (e.g., LTE-V2X, NR-V2X, BSM/CAM message families, and the standardized 802.11bd~\cite{ieee80211bd2023} alongside emerging 6G discussions) were not designed for arbitrary neural feature tensors. Near-term interoperability would therefore depend on explicit metadata schemas for shared features (including coordinate frames, timestamps, compression formats, uncertainty descriptors, and security credentials), even before any common latent-feature representation becomes standardized. Benchmarks such as V2X-Sim~\cite{li2022v2x} and OPV2V~\cite{xu2022opv2v} provide useful testbeds, but they should be complemented by protocol-level conformance tests, real-world channel measurements, and alignment with broader robotics middleware and UAV communication standards.

\subsection{Synthesis}

Trustworthy CI requires balancing safety, privacy, and utility, a triad that governs architectural decisions from the choice of fusion level to the design of communication protocols. The four axes surveyed above reveal a field in active tension: functional safety standards and certified robustness provide the regulatory foundation, privacy-preserving computation and differential privacy address the data exposure risks in multi-agent sharing, and policy-aware mechanisms around liability, fairness, and standardization define the socio-technical framework for deployment. Key open challenges include developing real-time Byzantine-robust aggregation with formal convergence guarantees, establishing interoperable neural payload standards across regulatory jurisdictions, and designing adaptive systems to navigate the safety-privacy-utility triad as conditions evolve.

\section{Open Challenges and Future Directions}\label{sec:open}

\begin{figure}[t]
	\centering
	\includegraphics[width=\linewidth]{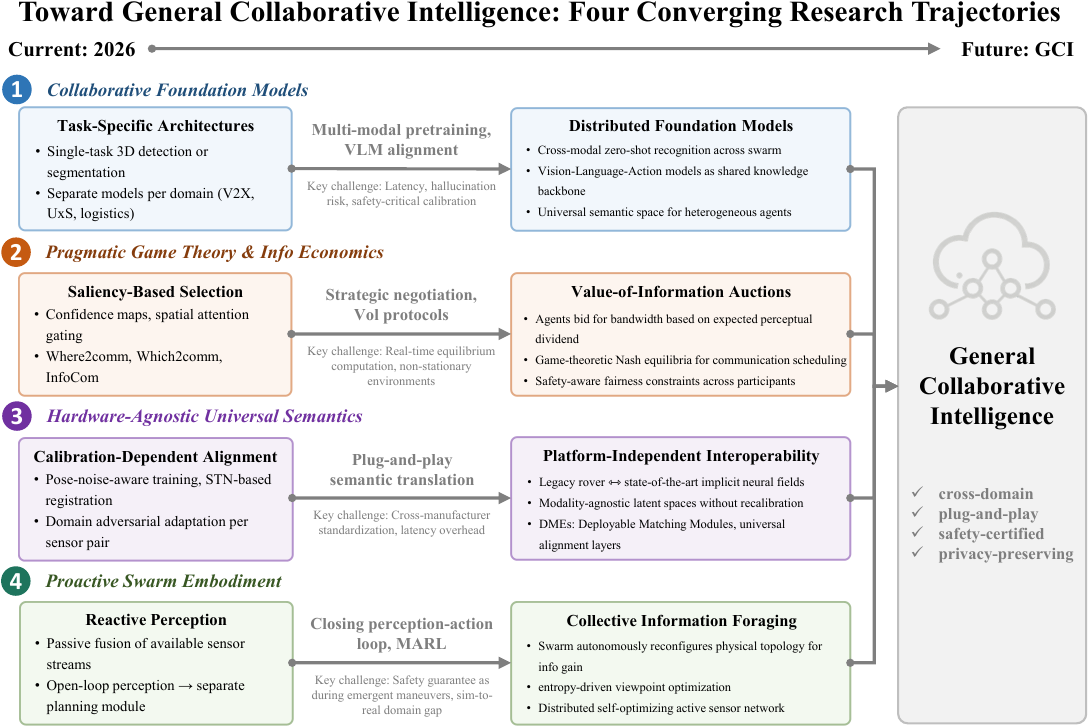}
	\caption{\textbf{Roadmap toward general collaborative intelligence.} Four converging research trajectories, i.e., collaborative foundation models, pragmatic game-theoretic communication, hardware-agnostic universal semantics, and proactive swarm embodiment, connect the current state of collaborative perception with the long-term vision of GCI.}
	\label{fig:gci_roadmap}
\end{figure}

The trajectory of GCI is currently traversing a fundamental inflection point: moving from task-specific fusion in controlled environments toward open-world cognitive synergy in human-centered environments. Beyond predefined datasets, technical feasibility meets socio-technical complexity. This section focuses on the future-facing pillars of open-world adaptation, pragmatic communication economics, semantic interoperability, embodied information foraging, and human-swarm cognitive integration. Figure~\ref{fig:gci_roadmap} summarizes these directions as four converging technical trajectories that link the current collaborative-perception paradigm to the longer-term goal of GCI. Before exploring these trajectories, this section first confronts the field's current limitations: the unresolved tensions and open problems that motivate the research agendas discussed below. The reflection below evaluates where current methods fall short on the five taxonomy dimensions and \Cone--\Cthree, and the future directions chart how to close those gaps.

\begin{table}[t]
	\centering
	\begin{threeparttable}
		\caption{\textbf{Taxonomy of open-world CI methods and cognitive synergy conditions.}}
		\label{tab:taxonomy_open}
		\renewcommand{\arraystretch}{1.5}
		\tiny
		\setlength{\tabcolsep}{4pt}
		\renewcommand{\tabularxcolumn}[1]{m{#1}}
		\sffamily
		\begin{tabularx}{\textwidth}{
				>{\raggedright\arraybackslash\hsize=1.3\hsize}X
				>{\centering\arraybackslash\hsize=0.7\hsize}X
				>{\centering\arraybackslash\hsize=0.8\hsize}X
				>{\raggedright\arraybackslash\hsize=1.4\hsize}X
				>{\centering\arraybackslash\hsize=0.9\hsize}X
				>{\raggedright\arraybackslash\hsize=1.5\hsize}X
				>{\centering\arraybackslash\hsize=0.7\hsize}X
				>{\centering\arraybackslash\hsize=0.7\hsize}X}
			\toprule
			\textbf{Method} & \textbf{Year} & \textbf{Stage} & \textbf{Comm.\ Paradigm} & \textbf{Architecture} & \textbf{Learning Strategy} & \textbf{Application} & \textbf{Synergy} \\
			\midrule
			Infinity-Driver~\cite{lai2024uncertainty} & \ygray{2024} & \InterBadge & On-demand & Decentr. & Contin. & \VTwoX & \ConeThree \\
			\rowcolor{gray!6} RACP~\cite{fang2025r} & \ygray{2025} & \InterBadge & IB-driven & Decentr. & Supervised + IB & \VTwoX & \ConeTwo \\
			Multi-VLA~\cite{shriram2025towards} & \ygray{2025} & \InterBadge & On-demand (VLM) & Decentr. & Pre-trained + VLM & \MultiDom & \ConeTwoThree \\
			\rowcolor{gray!6} MDrive~\cite{coscoy2026mdrive} & \ygray{2026} & \InterBadge & Periodic & Centr. & Supervised & \VTwoX & \ConeThree \\
			\bottomrule
		\end{tabularx}
		\smallskip
		\parbox{\linewidth}{\footnotesize \textit{Abbreviations:} VLM = Vision-Language Model, IB = Information Bottleneck, Contin.\ = Continual Learning, Decentr.\ = Decentralized, Centr.\ = Centralized. \textit{Synergy Conditions:} \Cone\ = Semantic Disambiguation, \Ctwo\ = Pragmatic Exchange, \Cthree\ = Proactive Foraging.}
	\end{threeparttable}
\end{table}

\subsection{Limitations and Critical Reflections}

While the preceding sections survey the substantial progress in collaborative perception, a candid assessment of the field's limitations is essential for guiding productive future research. We identify five interconnected problem areas that currently constrain the transition from laboratory benchmarks to robust real-world deployment.

\noindent\textbf{Benchmark Saturation and Evaluation Fragmentation.} The dominant benchmarks~\cite{xu2022opv2v,li2022v2x} drive significant progress but now exhibit early signs of saturation: recent methods cluster within a narrow band of a few percentage points on standard average precision (AP) metrics~\cite{zhang2025dsrc}, making it difficult to discern genuine advances from hyperparameter tuning and seed variance. Three concerns stand out. First, existing benchmarks involve only 2--5 agents, far below the tens to hundreds in realistic deployments (the scalability implications are discussed separately below). Second, the field relies overwhelmingly on detection-centric metrics (AP, NDS), neglecting communication efficiency, energy consumption, tail-latency, and graceful-degradation profiles. Third, inconsistent data splits, preprocessing, and hyperparameter reporting make cross-paper comparisons unreliable~\cite{guo2025autonomous}, and few works report seed variance.

\noindent\textbf{The Reproducibility Gap.} A significant fraction of methods in the collaborative perception literature lack publicly available implementations, and among those that do release code, the documentation and configuration files required to reproduce the exact reported results are often incomplete~\cite{pineau2021improving}. This problem is compounded by the inherent training instability of multi-agent systems: small changes in the communication schedule or the order in which agents contribute gradients can lead to substantially different convergence outcomes~\cite{henderson2018deep}. Unlike single-agent perception, where architectures and training recipes are extensively documented, multi-agent collaborative training introduces a combinatorial explosion of agent interaction dynamics that makes exact reproduction particularly fragile. Fields such as NLP and computer vision benefit substantially from initiatives like \textit{Papers with Code} and standardized leaderboards with locked test sets; collaborative perception has yet to establish comparable infrastructure.

\noindent\textbf{The Sim-to-Real Gulf.} With few exceptions, the empirical results reported in the collaborative perception literature are obtained within simulation environments or small-scale real-world V2X settings. While simulators such as CARLA and AirSim provide valuable controlled settings, the gap to physical deployment remains substantial and under-quantified~\cite{li2024s2r}. \textbf{Channel realism}: wireless channel models used in co-design training are predominantly AWGN or simple Rayleigh/Rician fading models; real mmWave and sub-6 GHz channels exhibit time-varying multipath, blockage effects, and spatially correlated shadowing that are rarely modeled~\cite{zhou2025ei}. \textbf{Sensor fidelity}: simulated LiDAR and camera outputs lack the noise characteristics, calibration drift, and environmental degradation (e.g., lens flare, LiDAR crosstalk in rain) that plague physical sensors. \textbf{Dataset realism}: recent real-world datasets have improved coverage, but most standard protocols still involve two or three cooperating agents, limited weather diversity, and constrained geographic settings. The field still lacks a large-scale, densely annotated, multi-agent real-world benchmark with more than five simultaneously cooperating agents and synchronized ground truth across diverse conditions. Until such benchmarks mature, the true performance of collaborative methods in deployment remains fundamentally uncertain. Recent work on closed-loop cooperative driving benchmarks (MDrive) provides early evidence that while multi-agent perception outperforms single-agent baselines, sharing perception features does not always translate to improved planning outcomes, which underscores the need for end-to-end evaluation beyond perception-only metrics~\cite{coscoy2026mdrive}.

\noindent\textbf{Negative Results and Failure Modes: When Collaboration Hurts.} The literature exhibits a strong positive-results bias: systems that successfully leverage collaboration are published, while conditions under which collaboration degrades performance are rarely documented. Several known failure modes deserve explicit attention. \textbf{Error propagation}: when a malfunctioning or adversarially compromised agent injects corrupted features into the fusion process, the collective world model can become \emph{worse} than what each agent would achieve independently: this phenomenon is documented in early work on Byzantine-resilient fusion~\cite{blanchard2017machine} but is rarely evaluated as a benchmark metric. \textbf{Correlated failures}: homogeneous agents with identical perception models and similar viewpoints are susceptible to correlated failures, where all agents miss the same occluded entity; collaboration does not help when the collective shares a common blind spot. \textbf{Communication-induced latency}: in time-critical scenarios such as highway emergency braking, the end-to-end latency introduced by feature encoding, transmission, and fusion can exceed the available decision budget, making collaborative perception actively detrimental compared to faster, albeit less informed, local-only inference~\cite{shenkut2024impact}. \textbf{Information cascades}: in decentralized architectures, a premature (and incorrect) semantic commitment by one agent can cascade through the swarm, with each agent reinforcing the error under the assumption that the collective belief is reliable. Systematic study of these failure modes, and of the conditions under which they dominate, remains an open research need.

\noindent\textbf{Scalability and the Agent-Churn Problem.} Most current methods assume a fixed set of collaborating agents with stable connectivity, a condition that rarely holds in practical deployments. Real-world swarms experience \textbf{agent churn}: vehicles enter and exit V2X range, UAVs are reassigned to different missions, and sensors are temporarily disabled by power constraints or environmental damage. The theoretical \(O(N^2)\) growth of pairwise feature interactions in transformer-based fusion architectures becomes computationally prohibitive as \(N\) grows to moderate numbers of agents on current-generation automotive-grade hardware~\cite{huang2024actformer, li2024collamamba}. Efficient approximations, such as learned sparsification of the attention graph~\cite{huang2024actformer} and hierarchical fusion topologies~\cite{li2024semantic}, are proposed but remain largely untested at the scales envisioned for smart-city or large-scale logistics deployment. Furthermore, the \textbf{cold-start problem} for newly joining agents (which lack the historical context shared by long-participating members) has received almost no attention in the literature~\cite{wang2025whales}.

Table~\ref{tab:open_challenges} consolidates the five problem areas identified above, assessing their severity, scope, and near-term tractability to guide future research prioritization.

\begin{table}[t]
	\centering
	\begin{threeparttable}
		\caption{\textbf{Open challenges in collaborative perception: severity, scope, and mitigation pathways.} Challenges are ordered approximately by their impact on the field's transition from laboratory benchmarks to robust real-world deployment.}
		\label{tab:open_challenges}
		\renewcommand{\arraystretch}{1.18}
		\tiny
		\setlength{\tabcolsep}{5pt}
		\renewcommand{\tabularxcolumn}[1]{m{#1}}
		\sffamily
		\begin{tabularx}{\textwidth}{
				>{\hsize=1.3\hsize\raggedright\arraybackslash}X 
				>{\hsize=0.5\hsize\centering\arraybackslash}X 
				>{\hsize=0.5\hsize\centering\arraybackslash}X 
				>{\hsize=0.5\hsize\centering\arraybackslash}X 
				>{\hsize=1.6\hsize\raggedright\arraybackslash}X 
				>{\hsize=1.6\hsize\raggedright\arraybackslash}X
			}
			\toprule
			
			\centering\textbf{Challenge} & 
			\textbf{Severity} & 
			\textbf{Scope} & 
			\textbf{Near-Term Solvability} & 
			\textbf{Representative Evidence} & 
			\centering\arraybackslash\textbf{Potential Mitigation Pathways} \\
			\midrule
			
			Benchmark Saturation & 
			High & 
			V2X domain & 
			Moderate & 
			AP clustering within few percentage points~\cite{zhang2025dsrc}; unreliable cross-paper comparisons~\cite{guo2025autonomous} & 
			Multi-dimensional metrics (accuracy--bandwidth--latency--energy); benchmarks with 10+ agents \\
			
			\cellcolor{gray!6} Reproducibility Gap & 
			\cellcolor{gray!6} High & 
			\cellcolor{gray!6} All domains & 
			\cellcolor{gray!6} Low & 
			\cellcolor{gray!6} Missing code and incomplete training configurations; multi-agent training instability~\cite{henderson2018deep} & 
			\cellcolor{gray!6} Standardized protocols~\cite{pineau2021improving}; containerized implementations; locked test-set leaderboards \\
			
			Sim-to-Real Gulf & 
			Critical & 
			All domains & 
			Low & 
			AWGN channel models vs.\ real mmWave~\cite{zhou2025ei}; sensor fidelity gaps~\cite{li2024s2r}; MDrive planning-perception gap~\cite{coscoy2026mdrive} & 
			Domain randomization; hardware-in-the-loop; large-scale real-world multi-agent benchmarks \\
			
			\cellcolor{gray!6} Under-Reported Failure Modes & 
			\cellcolor{gray!6} Moderate & 
			\cellcolor{gray!6} All domains & 
			\cellcolor{gray!6} Moderate & 
			\cellcolor{gray!6} LE/CE error taxonomy~\cite{guo2025autonomous}; Byzantine-robust fusion rarely benchmarked~\cite{blanchard2017machine}; latency-induced degradation~\cite{shenkut2024impact} & 
			\cellcolor{gray!6} Failure-analysis tracks; negative-result venues; standardized stress-test suites \\
			
			Scalability \& Agent Churn & 
			High & 
			Large-scale deploy. & 
			Low & 
			$O(N^2)$ communication growth~\cite{huang2024actformer}; cold-start problem unaddressed~\cite{wang2025whales} & 
			Hierarchical fusion~\cite{li2024semantic}; sparse attention~\cite{han2024agent}; incremental onboarding protocols \\
			\bottomrule
		\end{tabularx}
	\end{threeparttable}
\end{table}

\subsection{Toward General Collaborative Intelligence}

Current collaborative frameworks, while robust in high-fidelity simulations, are largely closed-world entities with architectures trained on static datasets that include predefined semantic categories. However, the physical world is characterized by a long-tail distribution of edge cases, where the safety-critical events are not encountered during training.

Since GCI is a forward-looking systems target, we state here what observation would count against it, lest every shortfall of current systems be taken as evidence for GCI's necessity rather than as evidence bearing on the concept's coherence. The GCI thesis rests on the premise that the local observation trap imposes a ceiling no single agent can shatter, so that pooling distributed observations yields capabilities unattainable alone under comparable sensing, communication, and compute. This thesis would be weakened if a single-agent foundation model, operating with zero collaboration bandwidth, were shown to match the perceptual accuracy and robustness of collaborative systems across diverse benchmarks, since the collaborative gain that justifies the entire enterprise would then be negligible. It would likewise be called into question if cross-domain shared world models proved fundamentally non-constructible, for instance if no common latent space could preserve task-relevant semantics across heterogeneous sensor modalities. We do not regard either outcome as likely given current evidence, but stating them turns GCI from an unfalsifiable aspiration into a claim that can, in principle, be tested.

\noindent\textbf{Collaborative Foundation Models and World-Model Priors.} Current collaborative frameworks are largely constrained to perception tasks, such as 3D object detection~\cite{xu2022opv2v} or semantic occupancy prediction~\cite{song2024collaborative}. The next generation of GCI will be powered by distributed foundation models that possess world-model priors, serving as shared knowledge backbones across the swarm~\cite{research0399}. By leveraging large-scale multi-modal pre-training (e.g., Vision-Language-Action models), a swarm can perform zero-shot identification of novel entities, from localized disaster debris to unconventional vehicles, through consensus-based latent reasoning~\cite{shriram2025towards}. In this paradigm, a drone observing a scene from above and a ground vehicle observing from the side do not merely match pixels; they exchange high-level semantic concepts: if one agent identifies a hazardous obstruction using its language-aligned encoder, the entire swarm inherits this semantic label without prior exposure to that specific visual geometry. Collaborative foundation models also enable heterogeneous agents (e.g., a LiDAR-equipped truck and a low-cost camera-based micro-drone) to map their disparate observations into a universal semantic space, ensuring that semantic understanding remains invariant when sensory modality shifts significantly. This direction, however, requires rigorous validation of latency, calibration, hallucination risks, and safety-critical failure modes in deployment before foundation-model-based collaboration can be entrusted with real-world safety decisions.

\noindent\textbf{LLM-Driven Multi-Agent Reasoning and Verbal Coordination.} Beyond serving as perception backbones, large language models open a qualitatively new coordination channel in which agents reason over a shared context and deliberate through structured natural-language exchange rather than implicit feature alignment. Multi-agent LLM collaboration, as studied in the agentic-AI literature~\cite{research0399}, decomposes a collective mission through explicit role specialization, assigning each agent a distinct function such as scout, planner, or safety verifier, and lets agents critique, vote on, and revise one another's proposals via multi-round debate. In a GCI setting this manifests as a cognitive layer atop the neural-communication stack: low-bandwidth semantic tokens summarizing each agent's local belief are aggregated into a shared prompt, and the resulting language-level consensus is projected back into per-agent control priors. Such verbal coordination is especially powerful for long-tail events absent from training data, where pretrained commonsense can interpolate plausible joint plans that no single agent could infer from pixels alone. The mechanism, however, is double-edged: ungrounded language generation can fabricate collectively adopted beliefs, so trustworthy deployment hinges on consensus protocols that cross-validate LLM outputs against sensor evidence and bound disagreement through calibrated uncertainty~\cite{hou2025driveagent}.

\noindent\textbf{Generative World Models as Counterfactual Simulators.} A second convergence is the recasting of world models from passive predictors into generative simulators that support counterfactual rollout. A swarm can hypothesize alternative collective maneuvers, such as ``what if the lead vehicle brakes instead of yielding?'', and evaluate each against a learned dynamics model before committing physical resources~\cite{fu2025generative}. This closes part of the sim-to-real gap diagnosed above: rather than relying on a fixed offline simulator, the swarm rehearses against its own continually updated, perception-grounded world model, exposing failure modes that only emerge under multi-agent coupling. When paired with LLM-based reasoning, the world model further supplies the hypothetical states over which language-level debate and verification operate, giving the cognitive layer a concrete, sensor-anchored substrate rather than free-form speculation.

\noindent\textbf{Consensus, Hallucination, and the Verification Gap.} The integration of foundation models into safety-critical collaboration exposes a verification gap that current GCI-Bench-style evaluation does not yet cover. LLM hallucinations, distribution shift under rare scenes, and the compounding of per-agent errors through consensus voting can produce confidently wrong collective decisions. Closing this gap will require (i) cross-agent evidence grounding that ties every language-level claim to a supporting sensory token, (ii) uncertainty-aware aggregation that down-weights agents whose perceptual confidence is low, and (iii) formal or runtime-verification fallbacks that trigger conservative behavioral envelopes when the collective belief diverges beyond a calibrated threshold~\cite{research0399}. Without such guarantees, the very opacity that makes foundation models expressive also makes them a liability in adversarial or safety-critical regimes.

\noindent\textbf{Continual Collaborative Learning and Decentralized Self-Healing.} Future swarms will need mechanisms for continual adaptation. As the collective navigates through new environments, it can perform decentralized self-labeling to update its internal representation without dense human annotation. A swarm can exploit viewpoint redundancy to perform real-time self-supervision~\cite{su2024makes}, where a high-confidence perspective from an un-occluded agent acts as the pseudo-label for the low-confidence, occluded views of its neighbors. As the system learns, it can distill new environmental knowledge into local weights or shared adapters~\cite{lai2024uncertainty}, improving robustness to seasonal changes, urban construction, or shifting traffic patterns.

\noindent\textbf{Hardware-Agnostic Interoperability and Universal Semantics.} Real-world deployments will be characterized by agent heterogeneity~\cite{xiang2023hm}. Achieving platform independence, where a legacy robotic rover seamlessly interoperates with a state-of-the-art agent utilizing collaborative implicit neural fields~\cite{zhao2024distributed}, remains a core challenge. Future research can advance modality-agnostic latent spaces and universal alignment layers~\cite{hu2024toward} that ensure semantic equivalence across disparate sensor resolutions and modalities, allowing heterogeneous agents to contribute to a unified geometric manifold without manual recalibration or task-specific retraining.

\noindent\textbf{The Economics of Pragmatic and Semantic Communication.} In contested or resource-scarce environments, communication is not merely a bandwidth problem but a strategic game of information theory~\cite{hu2026pragmatic}. The trade-off between perception fidelity and communication overhead is evolving into a complex economics of information. Moving beyond simple saliency-based selection~\cite{hu2022where2comm, yu2025which2comm}, and beyond the earlier who/when-to-communicate protocols that first made inter-agent communication a learnable, selective decision~\cite{liu2020who2com, liu2020when2com}, future GCI will utilize VoI-driven protocols in which agents quantify the utility of a feature patch before committing energy and bandwidth to transmit it. An agent evaluates whether sharing a given latent vector will meaningfully reduce the swarm's global entropy regarding a safety-critical object; if the information is redundant (e.g., observing a well-documented static building), it is suppressed~\cite{hu2022where2comm, chu2025occlusion}. By treating communication as a resource-constrained strategic game~\cite{hu2026pragmatic}, swarms negotiate over scarce bandwidth, energy, and compute, prioritizing the agent or viewpoint with the highest expected value of information while preserving safety constraints and fairness across participating agents~\cite{fang2025r}. This represents a deep integration of information theory and deep learning, maximizing the perceptual dividend per bit.

\noindent\textbf{Human-Swarm Cognitive Integration.} The long-term deployment of multi-agent systems is not in isolation but alongside human society, which necessitates a transition from black-box swarms to transparent cognitive partners. CI can be inherently explainable to its human stakeholders: in principle, a pedestrian in a smart city could query the collective belief and receive an interpretable response. Through generative synthesis (e.g., latent diffusion), the swarm can project its internal, high-dimensional world model back into a human-understandable visual or linguistic format~\cite{hou2025driveagent}, allowing operators to inspect the collective's spatial reasoning and resolving the alienation typically felt toward distributed AI systems. When an LLM backbone mediates this projection, the swarm can additionally narrate its reasoning in natural language: it can explain why it classified an ambiguous object as a hazard, identify which agent's evidence was decisive, and enumerate the alternative interpretations that were rejected. This turns the collective decision into an auditable chain of justification rather than a black-box verdict. By visualizing the swarm's confidence levels and uncertainty heatmaps, human operators can calibrate their trust~\cite{guo2024trust}, intervening when the swarm signals a high-entropy perceptual conflict that requires human common-sense intervention. In this symbiotic relationship, humans and swarms further leverage their complementary strengths: humans offload the cognitive burden of low-level spatial monitoring to the swarm~\cite{diaz2024externalized}, while the swarm leverages human social common-sense to resolve high-level semantic ambiguities.

\noindent\textbf{Collaborative Embodiment.}
A central challenge for GCI is the coupling of perception with proactive actuation at collective scale~\cite{coscoy2026mdrive}. Moving beyond passive observation, future swarms will exhibit emergent exploratory behaviors driven by the objective of collective entropy reduction: through proactive information foraging~\cite{nguyen2025csaot, zhou2025collaborative}, the swarm autonomously reconfigures its physical topology and viewing angles to actively resolve occlusions before they lead to perception failure. If the collective uncertainty in a specific sector rises, agents dynamically reposition to regain visual clarity, transforming the swarm into a distributed, self-optimizing active sensor. This represents a form of collective embodied intelligence that does not merely react to the world, but strategically probes it to maintain total situational clarity, pointing toward a future in which vehicles, infrastructure, and humans contribute to and draw from a shared, real-time global neural manifold.

\section{GCI-Bench: A Unified Evaluation Framework}\label{sec:gci-bench}

The critical reflection in Section~\ref{sec:open} diagnoses benchmark saturation and evaluation fragmentation as structural barriers to progress: current practice evaluates methods along narrowly defined axes (primarily detection AP on a single benchmark), ignores critical deployment dimensions such as communication cost and resilience, and reports results under inconsistent experimental conditions that preclude rigorous cross-method comparison. To address these limitations, we propose \textbf{GCI-Bench}, a unified evaluation framework that operationalizes the five-dimensional taxonomy of Figure~\ref{fig:taxonomy} into a quantitative scoring protocol. GCI-Bench is designed to be applied to any collaborative perception method without requiring access to its internal implementation and to produce a standardized ``GCI Scorecard'' that captures performance across the full design space.

\subsection{Five-Dimensional Evaluation System}

\begin{figure}[t]
\centering
\includegraphics[width=\textwidth]{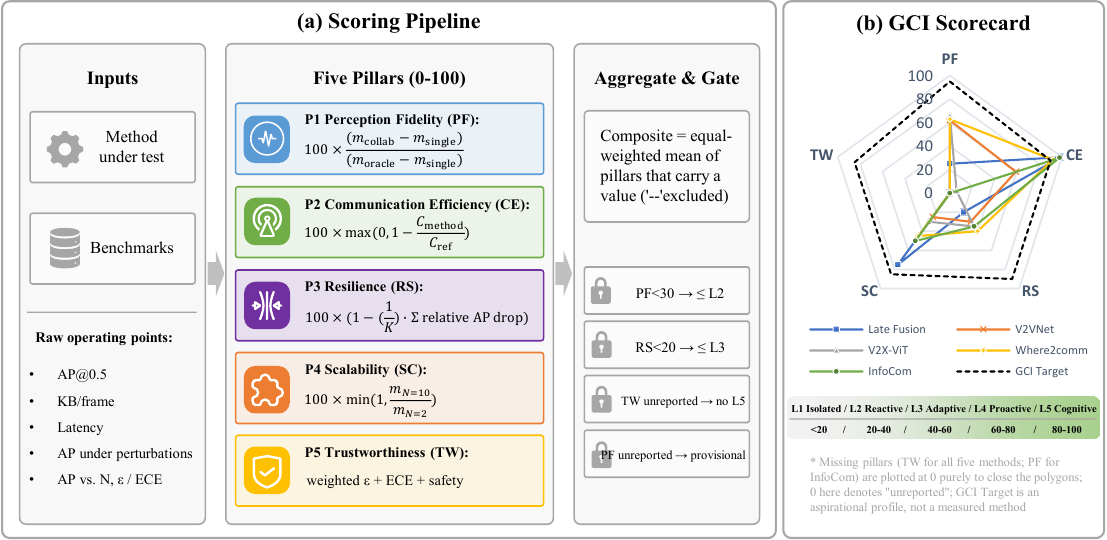}
\caption{\textbf{Overview of the GCI-Bench evaluation framework.} (a) Scoring pipeline: raw operating points reported by a method on standard benchmarks are normalized into the five pillars of Table~\ref{tab:gci_pillars}, each scored on a 0--100 scale against the disclosed single-agent and oracle anchors, and aggregated into a composite under the explicit missing-data rule and maturity caps. (b) Case-study scorecard: pillar profiles of the five representative methods of Table~\ref{tab:gci_case_study} together with the aspirational \textit{GCI Target} profile; unreported pillars are plotted at 0 only to close the polygons and denote missing rather than measured values.}
\label{fig:gci_bench}
\end{figure}

GCI-Bench defines five evaluation pillars, each scored on a 0--100 scale and normalized against a single-agent no-collaboration baseline (see Table~\ref{tab:gci_pillars} for the scoring protocol). The five pillars correspond to the taxonomy dimensions of Figure~\ref{fig:taxonomy}, ensuring that the evaluation framework is structurally aligned with the conceptual organization of the review. Figure~\ref{fig:gci_bench} gives an overview of the framework: the scoring pipeline that turns reported operating points into pillar scores, and the case-study scorecard that the pipeline produces.

\begin{table}[t]
	\centering
	\caption{\textbf{GCI-Bench five-pillar evaluation system.} Each pillar is scored on a 0--100 scale and normalized against a single-agent baseline. A weighted aggregate yields the GCI Composite Score.}
	\label{tab:gci_pillars}
	\renewcommand{\arraystretch}{1.20}
	\tiny
	\setlength{\tabcolsep}{5pt}
	\renewcommand{\tabularxcolumn}[1]{m{#1}}
	\sffamily
	\begin{tabularx}{\textwidth}{
			>{\hsize=0.5\hsize\raggedright\arraybackslash}X
			>{\hsize=1.0\hsize\raggedright\arraybackslash}X
			>{\hsize=1.0\hsize\raggedright\arraybackslash}X
			>{\hsize=2.0\hsize\raggedright\arraybackslash}X
			>{\hsize=0.5\hsize\raggedright\arraybackslash}X}
		\toprule
		\centering\textbf{Pillar} & \centering\textbf{What It Measures} & \centering\textbf{Primary Metrics} & \centering\textbf{Normalization} & \textbf{Benchmark Sources} \\
		\midrule
		P1. Perception Fidelity (PF) & Collaborative detection and tracking accuracy relative to single-agent baseline & AP@0.5, AP@0.7, NDS, AMOTA & $\mathrm{PF}=100\times(m_{\mathrm{collab}}-m_{\mathrm{single}})/(m_{\mathrm{oracle}}-m_{\mathrm{single}})$\newline \ygray{$m$ = primary detection metric; oracle = perfect-communication upper bound} & OPV2V~\cite{xu2022opv2v}, V2XSet~\cite{xu2022v2x}, DAIR-V2X~\cite{yu2022dair}, V2V4Real~\cite{xu2023v2v4real} \\
		\rowcolor{gray!6} P2. Communication Efficiency (CE) & Perceptual gain per unit of communication cost & Bandwidth (KB/frame), latency (ms), bits per AP gain & $\mathrm{CE}=100\times\max\bigl(0,\,1-C_{\mathrm{method}}/C_{\mathrm{ref}}\bigr)$, clamped to $[0,100]$\newline \ygray{$C$ = communication cost; $C_{\mathrm{ref}}$ = fixed reference cost, so $\mathrm{CE}=0$ at $C_{\mathrm{ref}}$} & All benchmarks with communication measurements \\
		P3. Resilience (RS) & Performance degradation under adverse conditions & AP drop under packet loss (10--50\%), pose noise ($\sigma=0.1$--$0.3$ m), adversarial perturbation & $\mathrm{RS}=100\times\Bigl(1-\frac{1}{K}\sum\nolimits_{k=1}^{K}\delta_k\Bigr)$\newline \ygray{$\delta_k$ = relative AP drop under perturbation type $k$; averaged over $K$ types} & OPV2V~\cite{xu2022opv2v}, V2X-Sim~\cite{li2022v2x}, custom stress-test suites~\cite{guo2025autonomous} \\
		\rowcolor{gray!6} P4. Scalability (SC) & Performance retention as the number of collaborating agents grows & AP at $N=2,5,10$; $O(N^2)$ communication growth factor & $\mathrm{SC}=100\times\min\bigl(1,\,m_{N=10}/m_{N=2}\bigr)$, clamped to $[0,100]$\newline \ygray{accuracy retention as the agent count grows from $N{=}2$ to $N{=}10$} & Benchmarks supporting 2--10 agents \\
		P5. Trustworthiness (TW) & Privacy preservation, uncertainty calibration, and safety compliance & Privacy budget $\varepsilon$, expected calibration error (ECE), ISO 26262/21448~\cite{ISO26262_2018, ISO21448_2019} compliance level & $\mathrm{TW}=w_{\varepsilon}\,\mathrm{TW}_{\varepsilon}+w_{\mathrm{ECE}}\,\mathrm{TW}_{\mathrm{ECE}}+w_{\mathrm{safety}}\,\mathrm{TW}_{\mathrm{safety}}$\newline \ygray{$w$ = domain-adjustable weights (summing to 1)} & DAIR-V2X~\cite{yu2022dair}, V2V4Real~\cite{xu2023v2v4real}, protocol-level tests \\
		\bottomrule
	\end{tabularx}
	\smallskip
	\parbox{\linewidth}{\footnotesize\textit{Note:} All scores are normalized such that the single-agent no-collaboration baseline scores 0 on every pillar. An oracle system with perfect communication and full omniscience would score 100. All pillar scores are clamped to $[0,100]$: negative scores (worse than no collaboration) and retention ratios above unity are truncated accordingly.}
\end{table}

\noindent\textbf{Pillar 1: Perception Fidelity (PF).} PF measures the core collaborative gain: the improvement in detection and tracking accuracy over a single-agent baseline. The oracle upper bound is defined as the performance when all agents share raw, lossless data under zero-latency conditions, representing the theoretical maximum of collaborative perception for a given architecture.

\noindent\textbf{Pillar 2: Communication Efficiency (CE).} CE penalizes methods that achieve strong accuracy through indiscriminate data broadcasting. The operational cost $C_{\text{method}}$ aggregates bandwidth and end-to-end latency; it is the engineering counterpart of the information-theoretic transmission rate $R(\mathbf{Z}_i)$ defined in Section~\ref{sec:foundations}, measuring what is actually spent on the channel rather than the mutual information conveyed. The linear normalization reaches 0 when the cost equals the reference $C_{\text{ref}}$ and uses a single reference cost fixed across benchmarks, preserving the cross-benchmark comparability that a per-benchmark calibration constant would undermine.

\noindent\textbf{Pillar 3: Resilience (RS).} RS measures average performance retention across a standardized suite of $K$ perturbation types (packet loss, pose noise, sensor degradation, adversarial perturbations), directly addressing the ``when collaboration hurts'' failure modes discussed in Section~\ref{sec:open}.

\noindent\textbf{Pillar 4: Scalability (SC).} SC measures performance retention as the agent count $N$ grows, penalizing methods whose $O(N^2)$ communication growth dominates; it operationalizes the agent-churn and cold-start concerns raised in Section~\ref{sec:open}. The ratio is clamped to $[0,100]$ so that the score cannot exceed 100 for methods whose accuracy improves as more agents join.

\noindent\textbf{Pillar 5: Trustworthiness (TW).} TW aggregates privacy preservation (differential privacy budget $\varepsilon$), uncertainty calibration (ECE), and safety compliance (ISO 26262/21448). The sub-weights $w$ are domain-adjustable: a privacy-sensitive smart-city deployment may raise $w_{\varepsilon}$, while a safety-critical V2X setting lets $w_{\text{safety}}$ dominate.

\subsection{Maturity Model and Scoring Protocol}

The five pillar scores are aggregated into a \textbf{GCI Composite Score} (weighted arithmetic mean, with domain-configurable weights defaulting to equal weighting) that maps onto a five-level maturity model, as defined in Table~\ref{tab:gci_maturity}.

\begin{table}[t]
	\centering
	\caption{\textbf{GCI maturity model.} Five capability levels defined by composite score ranges, with representative method archetypes.}
	\label{tab:gci_maturity}
	\renewcommand{\arraystretch}{1.20}
	\tiny
	\setlength{\tabcolsep}{6pt}
	\renewcommand{\tabularxcolumn}[1]{m{#1}}
	\sffamily
	\begin{tabularx}{\textwidth}{
			>{\hsize=0.2\hsize\centering\arraybackslash}X
			>{\hsize=0.4\hsize\centering\arraybackslash}X
			>{\hsize=0.7\hsize\centering\arraybackslash}X
			>{\hsize=2.0\hsize\raggedright\arraybackslash}X
			>{\hsize=1.7\hsize\raggedright\arraybackslash}X}
		\toprule
		\centering\textbf{Level} & \textbf{Designation} & \textbf{Score Range} & \textbf{Characteristics} & \textbf{Representative Archetype} \\
		\midrule
		L1 & Isolated & $<20$ & No effective collaboration; performance indistinguishable from or worse than single-agent baseline & Single-agent perception; naive late fusion that disregards neighbors' raw features \\
		\rowcolor{gray!6} L2 & Reactive & $20$--$40$ & Basic information sharing without adaptivity; fixed communication schedule & Broadcast-based intermediate fusion (V2VNet~\cite{wang2020v2vnet}, F-Cooper~\cite{chen2019f}) \\
		L3 & Adaptive & $40$--$60$ & Communication-aware collaboration with selective information transmission & Saliency-driven fusion (Where2comm, Which2comm) \\
		\rowcolor{gray!6} L4 & Proactive & $60$--$80$ & Embodied collaboration closing the perception-action loop; proactive information foraging & MARL-driven embodied systems (CSAOT, Plan2comm, CoPnP) \\
		L5 & Cognitive & $80$--$100$ & Trustworthy, self-optimizing, cross-domain distributed cognitive system; plug-and-play interoperability across heterogeneous fleets & Aspirational target for GCI; no current system fully achieves this level \\
		\bottomrule
	\end{tabularx}
\end{table}

The standardized scoring protocol operates as follows. For each method under evaluation, pillar scores are computed on each benchmark in the evaluation suite (currently OPV2V~\cite{xu2022opv2v}, V2XSet~\cite{xu2022v2x}, DAIR-V2X~\cite{yu2022dair}, and V2V4Real~\cite{xu2023v2v4real}). The per-benchmark scores are then aggregated via a weighted mean, with weights reflecting benchmark realism (real-world datasets receive higher weight than simulation-only). The aggregation follows an explicit missing-data rule: a pillar for which the underlying data are not reported is marked as unreported (``--'') and excluded from the composite, which is then the equal-weighted mean of the pillars that carry a value; composites averaged over different numbers of pillars are annotated and are not directly comparable. Each pillar score is reported with a 95\% confidence interval obtained through bootstrap resampling (1,000 iterations) over the test-set evaluation runs, where the resampling unit is the individual test-set scene (frame); such intervals are reported only under the unified re-evaluation protocol, because the per-scene runs required for resampling are unavailable when scores are computed from disclosed illustrative inputs, as in the case study of Section~\ref{sec:case_study}. The final output is a \textbf{GCI Scorecard}: a compact visual summary comprising a radar chart of the five pillar scores, the composite score with confidence bounds, and the assigned maturity level.

Three maturity caps refine the level assignment. A method with PF $<30$ cannot exceed L2, since a system that fails to outperform single-agent perception does not exhibit genuine collaboration. A method with RS $<20$ cannot exceed L3, since it does not retain accuracy under mild perturbation. A method whose TW pillar is unreported cannot be assigned L5, since trustworthiness is a necessary condition for cognitive maturity. A method whose Perception Fidelity pillar is unreported receives a provisional level, since fidelity anchors the composite and determines whether genuine collaboration is present at all; unreported complementary pillars such as Trustworthiness are instead excluded from the composite under the missing-data rule above.

\subsection{Case Study: Benchmarking Representative Methods}\label{sec:case_study}

To demonstrate the framework's operation, we apply GCI-Bench to five representative methods spanning the collaboration spectrum: Late Fusion~\cite{yu2022dair} (decision-level baseline), V2VNet~\cite{wang2020v2vnet} (broadcast intermediate fusion), V2X-ViT~\cite{xu2022v2x} (Transformer-based intermediate fusion), Where2comm~\cite{hu2022where2comm} (saliency-driven communication), and InfoCom~\cite{wei2026infocom} (information-bottleneck-driven extreme compression). All five are 3D object detection methods evaluated on OPV2V-class benchmarks. To make the demonstration auditable, we disclose below every constant and every raw input used to produce the scores of Table~\ref{tab:gci_case_study}, and we distinguish three kinds of cells: (i) scores computed from the disclosed illustrative operating points through the disclosed formulas (PF and CE); (ii) illustrative estimates that cannot yet be computed from a unified public suite (RS and SC); and (iii) pillars for which no underlying data are reported, marked as unreported (``--'') and excluded from the composite (TW for all five methods, and PF for InfoCom~\cite{wei2026infocom}).

For this demonstration we fix the following reference points, stated explicitly so that every score can be recomputed: for OPV2V (AP@0.5) we take the single-agent baseline $m_{\text{single}}=70$ and the perfect-communication oracle $m_{\text{oracle}}=90$; for V2XSet we take $m_{\text{single}}=75$ and $m_{\text{oracle}}=95$; and for the CE pillar we fix the reference cost at $C_{\text{ref}}=64$~KB/frame, the cost of broadcasting an uncompressed intermediate feature map, applied uniformly across benchmarks. These values are consistent with the no-collaboration and early-fusion operating points of the OPV2V benchmark (AP@0.5 $=0.679$ and $0.891$~\cite{xu2022opv2v}); they are illustrative rather than definitive, since unified reference values can only come from the controlled re-evaluation called for below.

To show that every computed cell of Table~\ref{tab:gci_case_study} is reproducible, we work one method end to end. Where2comm~\cite{hu2022where2comm} is assigned the illustrative operating point AP@0.5 $=82.5$ with an average uplink of $5$~KB/frame. Its Perception Fidelity is $\text{PF}=100\times\frac{82.5-70}{90-70}=62.5\approx63$. Its Communication Efficiency is $\text{CE}=100\times\max\bigl(0,1-\frac{5}{64}\bigr)=92.2\approx92$. Its Resilience (40) and Scalability (45) are illustrative estimates, because no unified public stress-test or agent-scaling suite yet reports the required perturbation and $N$-scaling curves; this absence is itself the evaluation gap identified in Section~\ref{sec:open}. Its Trustworthiness is unreported (``--''). The composite is the equal-weighted mean of the four valued pillars, $\frac{63+92+40+45}{4}=60$, corresponding to maturity level L3. The PF and CE scores of the remaining methods are produced by the same formulas from their assumed AP@0.5 and communication cost. Figure~\ref{fig:gci_bench}(b) overlays the resulting pillar profiles of all five case-study methods and of the \textit{GCI Target} row.

No comparable AP@0.5 operating point is available for InfoCom~\cite{wei2026infocom} under the demonstration's accounting, so its PF is marked unreported and its composite is averaged over the three remaining valued pillars (CE, RS, SC); its maturity level is therefore provisional. Late Fusion is assigned AP@0.5 $=75$, which yields $\text{PF}=25$ and triggers the PF~$<30$ maturity cap. The \textit{GCI Target} row is an aspirational profile scored on all five pillars and is visually separated from the five real methods; because its composite is averaged over five pillars whereas the real methods are averaged over four or fewer, the two are not directly comparable.

Published results illustrate how far current practice is from the unified accounting that the pillars presuppose. The OPV2V benchmark reports AP@0.5 $=0.858$ for late fusion and $0.908$ for attentive intermediate fusion on its default split, with no communication accounting of any kind~\cite{xu2022opv2v}. Where2comm re-evaluates the competing methods, itself included, under a dedicated protocol with a different AP scale and expresses communication as log$_2$-transformed bytes (5.67--22.71 across its operating points)~\cite{hu2022where2comm}. The two sources share neither an AP protocol nor a communication unit, so no pillar score can be computed for any method from published numbers alone; the case study therefore relies on disclosed illustrative inputs, and closing this accounting gap is precisely what the unified re-evaluation outlined below must deliver.

\begin{table}[t]
	\centering
	\caption{\textbf{GCI-Bench case study: pillar scores and reported operating points for five representative methods.} Normalised pillar scores (PF/CE/RS/SC/TW, 0--100) are derived from the disclosed illustrative operating points under the protocol of Table~\ref{tab:gci_pillars}; the three right-hand columns list the assumed raw operating points of the demonstration. Scores illustrate the framework rather than provide definitive rankings. ``--'' indicates that the original publication did not report data for that pillar.}
	\label{tab:gci_case_study}
	\renewcommand{\arraystretch}{1.15}
	\tiny
	\setlength{\tabcolsep}{4pt}
	\renewcommand{\tabularxcolumn}[1]{m{#1}}
	\sffamily
	\begin{tabularx}{\textwidth}{
			>{\hsize=1.0\hsize\raggedright\arraybackslash}X
			>{\hsize=0.6\hsize\centering\arraybackslash}X
			>{\hsize=0.6\hsize\centering\arraybackslash}X
			>{\hsize=0.4\hsize\centering\arraybackslash}X
			>{\hsize=0.4\hsize\centering\arraybackslash}X
			>{\hsize=0.4\hsize\centering\arraybackslash}X
			>{\hsize=0.4\hsize\centering\arraybackslash}X
			>{\hsize=0.4\hsize\centering\arraybackslash}X
			>{\hsize=0.6\hsize\centering\arraybackslash}X
			>{\hsize=0.5\hsize\centering\arraybackslash}X
			>{\hsize=0.5\hsize\centering\arraybackslash}X
			>{\hsize=0.5\hsize\centering\arraybackslash}X}
		\toprule
		\textbf{Method} & \textbf{Stage} & \textbf{Comm.} & \textbf{PF} & \textbf{CE} & \textbf{RS} & \textbf{SC} & \textbf{TW} & \textbf{Comp.} & \textbf{AP@.5} & \textbf{KB/f} & \textbf{Lat.} \\
		\midrule
		Late Fusion~\cite{yu2022dair} & Late & Metadata & 25 & 99 & 20 & 75 & -- & 55 (L2)$^{\dagger}$ & 75 & $<1$ & Low \\
		\rowcolor{gray!6} V2VNet~\cite{wang2020v2vnet} & Intermed. & Broadcast & 62 & 59 & 30 & 25 & -- & 44 (L3) & 82.4 & 26 & Med \\
		V2X-ViT~\cite{xu2022v2x} & Intermed. & Broadcast & 66 & 6 & 35 & 30 & -- & 34 (L2) & 88.2$^{\ddagger}$ & 60 & Med \\
		\rowcolor{gray!6} Where2comm~\cite{hu2022where2comm} & Intermed. & Saliency & 63 & 92 & 40 & 45 & -- & 60 (L3) & 82.5 & 5 & Med \\
		InfoCom~\cite{wei2026infocom} & Intermed. & IB-pruned & -- & 98 & 35 & 50 & -- & 61 (prov.)$^{\S}$ & -- & $\sim 1$ & Med \\
		\midrule
		\rowcolor{gray!6} \textit{GCI Target} & -- & -- & 95 & 90 & 90 & 85 & 85 & 89 (L5) & $>95$ & $<5$ & Low \\
		\bottomrule
	\end{tabularx}
	\smallskip
	\parbox{\linewidth}{\tiny\textit{Note:} PF = Perception Fidelity, CE = Communication Efficiency, RS = Resilience, SC = Scalability, TW = Trustworthiness, Comp.\ = composite, Lat.\ = latency class. AP@.5 = AP@0.5 IoU on the method's native benchmark (OPV2V unless otherwise noted); KB/f = average uplink per agent per frame; `--' = not reported in the original publication. PF and CE are computed from the disclosed formulas of Table~\ref{tab:gci_pillars} using the reference constants stated in the text ($m_{\text{single}}$, $m_{\text{oracle}}$, $C_{\text{ref}}$) together with each method's assumed AP@0.5 and communication cost; RS and SC are illustrative estimates rather than values computed from a unified public suite (see text); TW is unreported for all five methods. Per the missing-data rule, each composite is the equal-weighted mean of the pillars carrying a value. $^{\ddagger}$V2X-ViT's AP@0.5 is reported on its native V2XSet benchmark rather than OPV2V, and its PF and CE use the corresponding V2XSet reference points. InfoCom's PF is marked `--' because no comparable AP operating point is available under the demonstration's accounting; its composite is averaged over the three valued pillars (CE, RS, SC) and its maturity level is provisional ($^{\S}$). The raw operating points are disclosed illustrative values spanning the stage--communication design space rather than transcribed measurements; published sources share neither an AP protocol nor a communication unit (see text), so the scores should not be read as a controlled cross-method ranking. The \textit{GCI Target} row is an aspirational profile scored on all five pillars and is not directly comparable to the real-method composites, which are averaged over four or fewer pillars. $^{\dagger}$A maturity cap applies: methods with PF $<$ 30 cannot exceed L2 regardless of composite score, since a system that fails to outperform single-agent perception is not considered to exhibit genuine collaboration.}
\end{table}

Several patterns emerge from the case study. First, communication-efficient methods achieve higher composite scores than broadcast-based methods despite slightly lower raw perception fidelity (Where2comm, InfoCom), because the CE pillar rewards bandwidth discipline. Second, no current method reports comprehensive data across all five pillars: fidelity and cost are reported widely enough to compute PF and CE, but RS and SC can only be estimated illustratively and TW is unreported for every method, confirming the evaluation gap identified in Section~\ref{sec:open}. Third, the maturity levels suggest that the field currently operates predominantly at L2--L3 (reactive to adaptive collaboration), with L4 (proactive embodied collaboration) demonstrated only in research prototypes and L5 (cognitive) remaining aspirational; InfoCom's level is provisional because its PF pillar is unreported. The \textit{GCI Target} row in Table~\ref{tab:gci_case_study} defines the aspirational performance profile that a mature GCI system would need to achieve.

\subsection{Limitations of the Framework and Community Adoption Path}

GCI-Bench is a conceptual framework and its practical deployment faces several challenges. First, the pillar scores computed in the case study are derived from heterogeneous experimental conditions across original publications (different data splits, hyperparameters, hardware platforms); definitive cross-method comparison requires unified re-evaluation under controlled conditions. Second, the pillar weights and the maturity-level thresholds are initial proposals that should be refined through community consensus, for instance via a dedicated workshop or challenge track at a major venue. Third, the trustworthiness pillar (TW) is the least developed, as most current methods do not report privacy budgets, calibration errors, or safety compliance data; populating this pillar will require the community to adopt standardized reporting practices for these dimensions. Fourth, the oracle upper bound used for PF normalization is architecture-dependent and may not be attainable for all method classes. Fifth, the three cognitive synergy conditions (C1--C3) are formulated as operational design desiderata rather than a formal theorem; establishing their near-necessity, for instance via an information-theoretic impossibility result showing that no proper subset of the conditions can achieve cognitive synergy, remains an open theoretical question beyond the scope of this review.

We are explicit about the sense in which the framework is and is not validated. The C1--C3 badge assignments, although now governed by the adjudication rule of Section~\ref{sec:foundations}, remain author judgments and have not been independently re-labeled; the case-study scores are computed from disclosed illustrative operating points rather than measured under a unified protocol; and the reference constants used for the demonstration ($m_{\text{single}}$, $m_{\text{oracle}}$, $C_{\text{ref}}$) are illustrative rather than unified. Consequently, the case study demonstrates internal consistency and reproducibility of the scoring chain, not the external validity of any particular ranking. An external re-labeling of the badge assignments and a controlled re-evaluation of the case-study methods under the unified protocol would strengthen validity, and we leave both to the community effort outlined below.

A natural alternative to GCI-Bench is to strengthen the reporting norms of existing leaderboards, for example locking test splits and mandating bandwidth and latency columns on OPV2V and DAIR-V2X. We regard such norms as complementary rather than sufficient: they improve comparability \emph{within} a single benchmark, but they cannot express the dimensions that differentiate collaborative systems \emph{across} benchmarks and domains, since a bandwidth column captures neither a resilience degradation curve nor a scalability slope nor a trustworthiness profile, and locked splits do not penalize a method that achieves accuracy through profligate broadcasting. GCI-Bench is intended to be adopted \emph{alongside} existing leaderboards, and we acknowledge the adoption cost it asks of the community, namely shared compute for unified re-evaluation, converged perturbation suites, and reporting practices that do not yet exist; the staged path below therefore begins with reporting norms and evolves toward the full apparatus. Finally, the framework's relevance does not depend on collaboration always winning: if single-agent foundation models shrink the collaborative gain that PF measures, the remaining pillars still quantify what a strong single agent cannot obtain alone, namely bandwidth discipline (CE), degradation behavior (RS), scaling retention (SC), and trustworthiness (TW), so the framework degrades gracefully rather than becoming irrelevant.

We recommend three steps toward community adoption. (1) \textbf{Benchmark integration}: existing benchmarks (OPV2V~\cite{xu2022opv2v}, V2X-Sim~\cite{li2022v2x}, DAIR-V2X~\cite{yu2022dair}, V2V4Real~\cite{xu2023v2v4real}) should adopt the five-pillar reporting template as a supplementary evaluation protocol alongside their current metrics. (2) \textbf{Standardized stress-test suites}: the community should converge on a shared set of perturbation configurations for the RS pillar (packet loss rates, pose noise levels, adversarial attack types) to ensure comparability. (3) \textbf{Annual GCI-Bench challenge}: a community-organized challenge modeled on established benchmarks (e.g., nuScenes tracking challenge, Waymo Open Dataset challenge) would provide a neutral platform for unified re-evaluation and incentivize comprehensive reporting across all five pillars. The GCI-Bench framework, by making the evaluation dimensions of Figure~\ref{fig:taxonomy} quantitatively operational, provides the scaffolding for such a community effort. Looking forward, the framework must also accommodate emerging foundation-model-driven collaborative systems discussed in Section~\ref{sec:open}. The PF and CE pillars extend naturally to VLM/VLA-based agents, but two gaps remain: (i) the RS pillar currently perturbs only the perceptual channel, whereas foundation-model collaboration introduces a new failure surface, i.e., LLM hallucination and ungrounded consensus, that calls for a dedicated ``cognitive resilience'' sub-metric measuring belief drift under adversarial or out-of-distribution queries; and (ii) the TW pillar should be augmented with a verifiability component that scores whether a collective decision can be traced to supporting sensory evidence rather than fabricated reasoning. Extending GCI-Bench along these axes is a natural prerequisite for scoring methods toward L5 (Cognitive) maturity.

\section{Conclusion}\label{sec:conclusion}

This review has charted the evolution of multi-agent collaborative intelligence from deterministic data fusion toward cognitive synergy, organized around a five-dimensional taxonomy and three operational synergy conditions (C1--C3) that together span the design and evaluation space of the field. The GCI-Bench framework operationalizes this taxonomy into a multi-pillar scoring protocol, and the critical reflection on benchmark saturation, reproducibility, the sim-to-real gulf, and failure modes where collaboration degrades performance offers a counterweight to the positive-results bias prevalent in the literature. The technical arc traced across architectures, neural-communication co-design, embodied synergy, resilience, domains, and trust points toward a convergence of collaborative foundation models, pragmatic communication, hardware-agnostic semantics, and proactive swarm embodiment, with the software-defined collaborative network providing the substrate through which the C1--C3 conditions can be jointly realized.
Collaborative intelligence is more than an enhancement of robotic sensing: it enables distributed systems that perceive, reason, and act with resilience exceeding the sum of their parts. As these capabilities mature and integrate into smart cities, logistics networks, and environmental monitoring, the boundary between ego-centric perception and collective cognition will continue to blur, and the central challenge for the coming decade will be turning these converging trajectories into cohesive, deployable general collaborative intelligence.

\section*{Acknowledgments}
This work was supported by the National Natural Science Foundation of China under Grant 62401447, and by the Natural Science Basic Research Program of Shaanxi under Grants 2026JC-YXQN-127 and 2026JC-YXQN-124.

\printbibliography

\end{document}